\documentclass{naturep}

\usepackage{amssymb}
\usepackage{amsmath}
\usepackage{graphicx}
\usepackage{algorithmicx}
\usepackage{algorithm}
\usepackage{dsfont}
\usepackage[noend]{algpseudocode}
\usepackage{bbm}
\usepackage[displaymath]{lineno}
\usepackage[margin=0.6in]{geometry}
\usepackage{ragged2e}
\usepackage{setspace}
\usepackage{hyperref}
\usepackage{caption}
\usepackage{subcaption}
\usepackage{anyfontsize}
\usepackage{setspace}
\usepackage{float}
\usepackage{bm}
\usepackage{multirow}
\usepackage{booktabs}
\usepackage[dvipsnames]{xcolor}

\makeatletter
\let\saved@includegraphics\includegraphics
\AtBeginDocument{\let\includegraphics\saved@includegraphics}

\makeatother

\title{\begin{flushleft}{\begin{spacing}{1}Fast and Scalable Image Search For Histology\end{spacing}}\end{flushleft}}

\begin{document}
\maketitle
\begin{spacing}{1.8}
\vspace{-15mm}
\noindent Chengkuan Chen$^{1,2,3}$, Ming Y. Lu$^{1,2,3,5}$, Drew F. K. Williamson$^{1,2,5}$, Tiffany Y. Chen$^{1,2,5}$,  Andrew J. Schaumberg$^{1}$, and Faisal Mahmood$^{*1,2,3}$\\
\begin{affiliations}
 \item Department of Pathology, Brigham and Women's Hospital, Harvard Medical School, Boston, MA
 \item Cancer Program, Broad Institute of Harvard and MIT, Cambridge, MA 
 \item Cancer Data Science, Dana-Farber Cancer Institute, Boston, MA
 \item Contributed Equally
 \end{affiliations}
 
 \noindent\textbf{Code / Package:} http://github.com/mahmoodlab/FISH\\
% \textbf{Interactive Demo:} http://clam.mahmoodlab.org\\
 
\end{spacing}
\begin{spacing}{1.4}
\noindent\textbf{*Correspondence:}\\ 
Faisal Mahmood \\
60 Fenwood Road, Hale Building for Transformative Medicine\\
Brigham and Women's Hospital, Harvard Medical School\\
Boston, MA 02445\\
faisalmahmood@bwh.harvard.edu
\end{spacing}

\newpage
%\linenumbers
\noindent\textbf{\large{Abstract}}
\vspace{-5mm}
\begin{spacing}{1.3}
\noindent The expanding adoption of digital pathology has enabled the curation of large repositories of histology whole slide images (WSIs), which contain a wealth of information. Similar pathology image search offers the opportunity to comb through large historical repositories of gigapixel WSIs to identify cases with similar morphological features and can be particularly useful for diagnosing rare diseases, identifying similar cases for predicting prognosis, treatment outcomes and potential clinical trial success. A critical challenge in developing a WSI search and retrieval system is scalability, which is uniquely challenging given the need to search a growing number of slides that each can consist of billions of pixels and are several gigabytes in size. Such systems are typically slow and retrieval speed often scales with the size of the repository they search through, making their clinical adoption tedious and are not feasible for repositories that are constantly growing. Here we present Fast Image Search for Histopathology (FISH), a histology image search pipeline that is infinitely scalable and achieves constant search speed that is independent of the image database size, while being interpretable and without requiring detailed annotations. FISH uses self-supervised deep learning to encode meaningful representations from WSIs and a Van Emde Boas tree for fast search, followed by an uncertainty-based ranking algorithm to retrieve similar WSIs. We evaluated FISH on multiple tasks and datasets with over 22,000 patient cases spanning 56 disease subtypes. We additionally demonstrate that FISH can be used to assist with the diagnosis of rare cancer types where sufficient cases may not be available to train traditional supervised deep models. FISH is available as an easy-to-use, open source software package (https://github.com/mahmoodlab/FISH). 
\end{spacing}
%\end{abstract}

\newpage
\begin{spacing}{1.42}
The increasing availability of technologies allowing for the routine creation of high resolution whole slide images (WSIs) has triggered tremendous excitement for the field of digital pathology. Where rich and massive image data generated and digested by pathologists was once locked in glass slides, whole slide imaging systems now allow pathologists and researchers to access that data without a microscope at hand. Studies demonstrating non-inferiority of WSIs\cite{snead2016validation,mukhopadhyay2018whole,azam2020diagnostic} and approvals by the FDA allowing primary diagnosis to be performed on WSIs mean that pathologists can now adopt these systems for clinical use. However, as institutions scan and store increasing numbers of images, they often turn to WSI storage and retrieval paradigms identical to that of their glass slides--large repositories of data searchable by patient identifiers, case number, date of procedure, pathology report, \textit{etc.}, without leveraging the digital content of the images themselves. 
\vspace{-4mm}

The revolution of artificial intelligence\cite{lecun2015deep, esteva2019guide} (\textit{e.g.}, deep learning) in recent years has shown potential in various tasks in pathology that range from disease diagnosis, prognosis, and integrative multi-omic analysis\cite{esteva2017dermatologist, bera2019artificial, niazi2019digital, lu2021data, chen2020pathomic, lu2021toad}. A critical challenge that hinders the progress in large scale, efficient adoption of histology whole slide images is the scalability of the search system, which is uniquely challenging for WSI retrieval systems\cite{komura2018machine} (as compared to other image databases) as they need to efficiently search a growing number of slides that each can consist of billions of pixels and are gigabytes in size--data that is much too big for traditional image search and retrieval solutions such as Google's Reverse Image Search.
\end{spacing}
% \begin{figure*}[b!]
% \centering
% \includegraphics[width=0.8\textwidth]{figs/FIG0_1.jpg}
% \caption{\textbf{The conceptual diagram of Whole Slide Images (WSIs) retrieval.}. Given a query WSI, the retrieval system should return the top $K$ similar whole slides and their associated report such as diagnosis.}
% \label{fig_0}
% \end{figure*}
\vspace{-4mm}
\begin{spacing}{1.42}
As such, most approaches split the WSIs into image patches and either focus on patch or region of interest (ROI) retrieval that is tailored to specific applications\cite{qi2014content,zhang2014towards, sridhar2015content,kwak2016automated,sparks2016out,jiang2016scalable,shi2017supervised,komura2018luigi,schaer2019deep,ma2016breast, zheng2018histopathological, akakin2012content}. These implementations often need expert pathologists to exhaustively delineate the ROIs, making the system difficult to scale due to the lack of pathologists or pathologist time. Though promising patch retrieval results in a recent paper have been demonstrated without manual labels\cite{hegde47822}, the desired results consume significant computing resources in the cloud to achieve and its search speed scales with the size of the dataset. In a line of recent literature\cite{kalra2020pan, kalra2020yottixel}, the authors propose a method that converts whole slide images into a set of barcodes that enables  search in a large database with low storage cost. However, the disadvantage of the proposed method is slow search speed on larger datasets due to $O(n\log(n))$ computational complexity. Additionally, the reported results show weakness when the number of slides is skewed in the anatomical site, which is commonly seen in real world histology datasets. Some recent works attempt to create improved feature representation for whole slide images by creating permutation invariant embeddings\cite{hemati2021cnn} or fine-tuning pretrained networks on data with morphological information\cite{riasatian2021fine}. These studies can not generate interpretable results and lack evaluation on large and diverse cohorts, which makes it difficult to assess their clinical benefits. Scalibility to large histological datasets with real world biases in disease types is crucial for a practical search engine in histology.
\vspace{-4mm}

Here, to overcome these challanges, we propose Fast Image Search for Histopathology (FISH) as a new search pipeline that addresses the issues summarized above. FISH guarantees $O(1)$ (constant time) theoretical query speed by representing a WSI as a set of integers and a binary code, and does not require any ROI annotations. We evaluate FISH on two main tasks: first, performance on disease subtype retrieval from a fixed anatomic site as tested on three independent cohorts, specifically, the primary diagnostic slides in The Cancer Genome Atlas (TCGA), Clinical Proteomic Tumor Analysis Consortium (CPTAC) and in-house Brigham Women's Hospital data; and second, performance on retrieving slides from the same anatomic site as the query, which we evaluate using the TCGA dataset. In total, we used 22,385 diagnostic whole slide images across 13 anatomic sites and 56 disease subtypes. Finally, we also demonstrate that FISH can perform patch-level search in $O(1)$ (even though the pipeline is not specifically designed for the task), which makes it the first system that can handle both slide-level and patch-level query in histology image retrieval.

%  Although we always know where the slides are scanned from, we believe this task is helpful in archiving slides in past several decades where site information is not available. This task also serves a benchmark to test the search efficiency when we have a large number of slides.
\vspace{-4mm}

 FISH  uses Vector Quantised-Variational AutoEncoder (VQ-VAE)\cite{oord2017neural} trained on a large dataset in an unsupervised manner and leverages the discrete latent code to create integer indices for automatically selected patches in a WSI. With the integer representation of a slide, we can benefit from the $O(\log\log(M))$ search speed provided by Van Emde Boas tree (vEB tree) where $M$ is a fixed constant in our pipeline. Furthermore, we also show that FISH can provide interpretable results during the ranking stage, which is a desirable property for any medical application to help the decision process. Finally, we make the source code of FISH open access (link:https://github.com/mahmoodlab/FISH) for future studies.
\end{spacing}

\begin{figure*}
\vspace{-12mm}
\includegraphics[width=\textwidth]{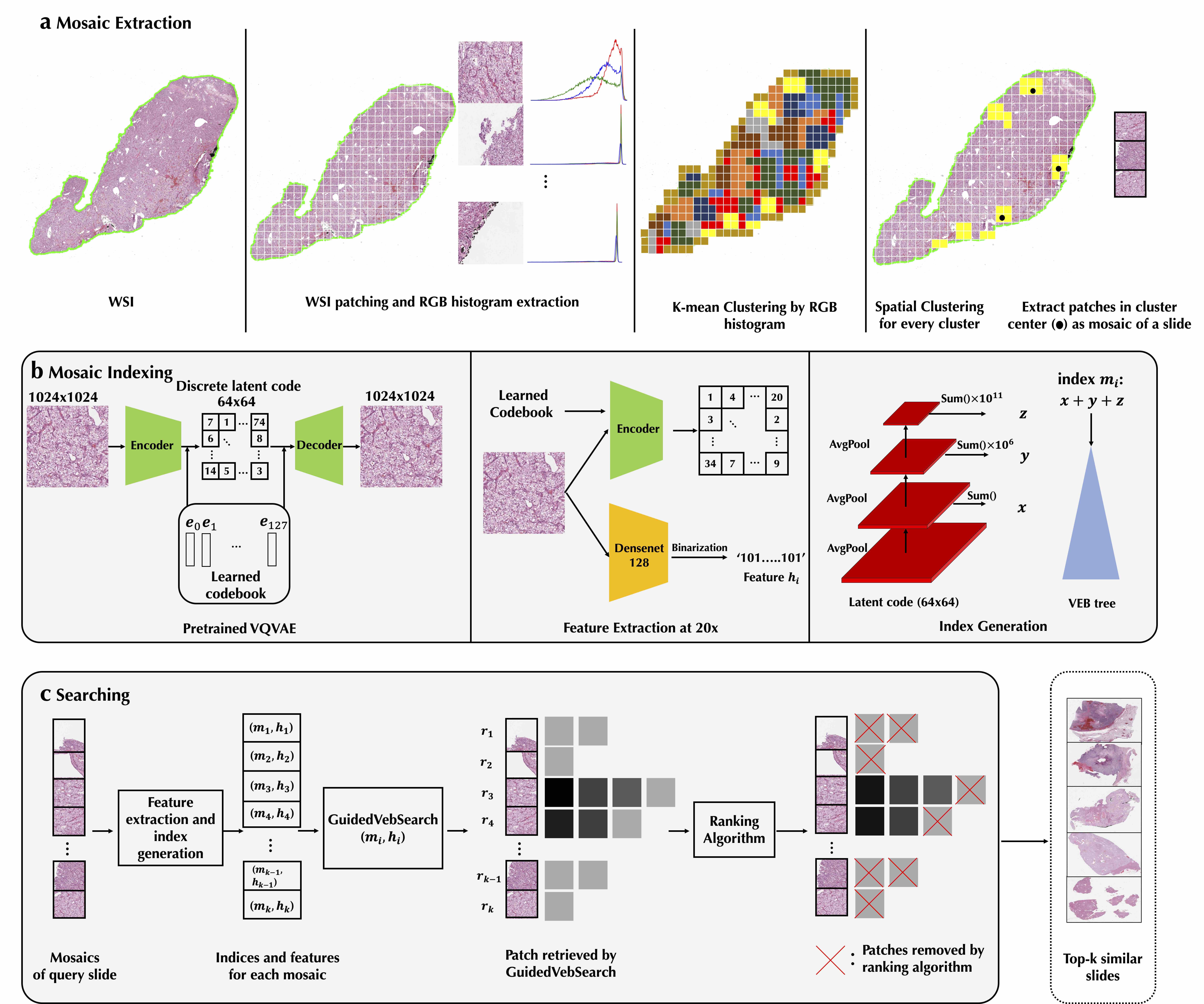}
%\vspace{-3mm}
\caption{\textbf{Overview of the FISH pipeline.} \textbf{a} After segmentation, we extract RGB histogram as features from each patch for K-mean clustering. To get the final mosaics of the given slide, in each cluster generated from the previous stage (\textit{e.g.}, the yellow cluster shown in the figure.), we use the coordinates of the patch as features then apply K-mean clustering to them (\textit{i.e.}, spatial clustering) and extract patches from the cluster center coordinates (black dots) as mosaics for a slide. \textbf{b} We pretrained a VQ-VAE on TCGA and use its encoder and codebook for feature extraction. In the feature extraction stage, the first extractor is the encoder in VQ-VAE used to get a discrete latent code and the second one is Densenet used to get the binarized texture feature of the patch. Finally, we feed the latent code into another pipeline composed of a series of pool, shift, and summation to get an integer index for the patch then use the vEB tree to construct the index structure for search. \textbf{c} For a given query slide preprocessed as mosaics representation, we feed them into the feature extractor to get the corresponding index and binarized texture features of each mosaic, then apply our search and ranking algorithm to filter out the candidate mosaics (\textit{i.e.}, mosaics with patches remained). Since each patch in the candidate mosaics contains a slide name where the patch is from, we visit the patches in the candidate mosaics
in the ascending order of hamming distance (from dark to light) then use the corresponding slides name to return top-K similar slides. }
\label{fig_1}
\end{figure*}
\begin{spacing}{1.45}
\vspace{-6mm}
\section*{FISH: A scalable search engine for histology images}
\vspace{-4mm}
FISH is a deep learning based histology image retrieval method that combines VQ-VAE\cite{oord2017neural} and vEB tree to achieve $O(1)$ search performance with low storage cost, while also supporting patch-level retrieval and human-interpretability. To achieve this performance, we represent each slide image as a set of integers and binary codes for efficient storage and encode the integers into a vEB tree for fast search. The overview of FISH pipeline is shown in \textbf{Figure~\ref{fig_1}}.
\vspace{-4mm}

FISH begins by distilling a mosaic representation of a given slide\cite{kalra2020yottixel}. To select the patches used for representing the slide we used two-stage K-means clustering. Specifically, we first apply K-means clustering on the RGB features extracted from patches at 5$\times$, followed by K-means clustering on the coordinates of patches at 20$\times$ within each initial cluster. We extract image patches corresponding the coordinates of the set of final cluster centers and use them as a mosaic representation of the given slide. To convert the mosaics into a set of integers and binary code (\textbf{Figure~\ref{fig_1}b}), we pre-train a VQ-VAE, which is a variant of a Variational Autoencoder\cite{kingma2013auto} that gives the input a discrete latent code from a codebook learned on the TCGA slides at 20$\times$.  We use the encoder of the pretrained VQ-VAE along with the learned codebook to encode the patches at 20$\times$ and extract mosaic features by using a Densenet\cite{huang2017densely} model and a binarization algorithm. The last step is to convert the discrete latent codes into integers to store the mosaics in the vEB tree. We feed the latent codes of the mosaics into a pipeline composed of a series of average pooling (AvgPools), summation, and shift operations. The intuition behind this pipeline is to summarize the information in each scale via summation then store it into a different range of digits in a integer.
\vspace{-4mm}

During searching (\textbf{Figure~\ref{fig_1}c}), we extract the features of the preprocessed mosaics of the query whole slide image and then apply the proposed Guided Search Algorithm (GSA) to find the most similar results of each mosaic. The design principle of GSA is to find a fixed number of the nearest neighbors using the vEB and only select the neighbors whose hamming distances from the query mosaic are lower than a certain threshold $\theta_{h}$. Since we only look for a fixed number of neighbors and the search time of the vEB to find neighbors is $O(1)$, the time complexity of FISH search is $O(1)$. The search result of each mosaic is a list of patches. Each patch contains metadata that documents the name of the slide where the patch is located, the diagnosis of the slide, and hamming distance between the patch and the query mosaic. Once each mosaic gets its search results, our ranking algorithm ranks the candidate mosaics used to retrieve the final top-K similar slide. We collect all slides that appear in the search results from the candidate mosaics and sort them based on hamming distance in ascending order to return the top-K similar slides.
\vspace{-4mm}

In the next sections, we demonstrate performance in four areas: (1) disease subtype retrieval in the fixed anatomic site in public cohorts (TCGA, CPTAC), (2) disease subtype retrieval in the fixed anatomic site in independent cohorts (BWH in-house data) to test generalizability, (3) disease anatomic site retrieval, and (4) speed and interpretability. In addition, we also report that our system can handle patch level retrieval with $O(1)$ search performance on Kather100k\cite{kather2019predicting} and in-house prostate data though the system is not specifically designed for this task.
\end{spacing}
\vspace{-8mm}
\section*{\large{Results}}
\begin{figure*}[t!]
    \centering
    \includegraphics[width=0.88\textwidth]{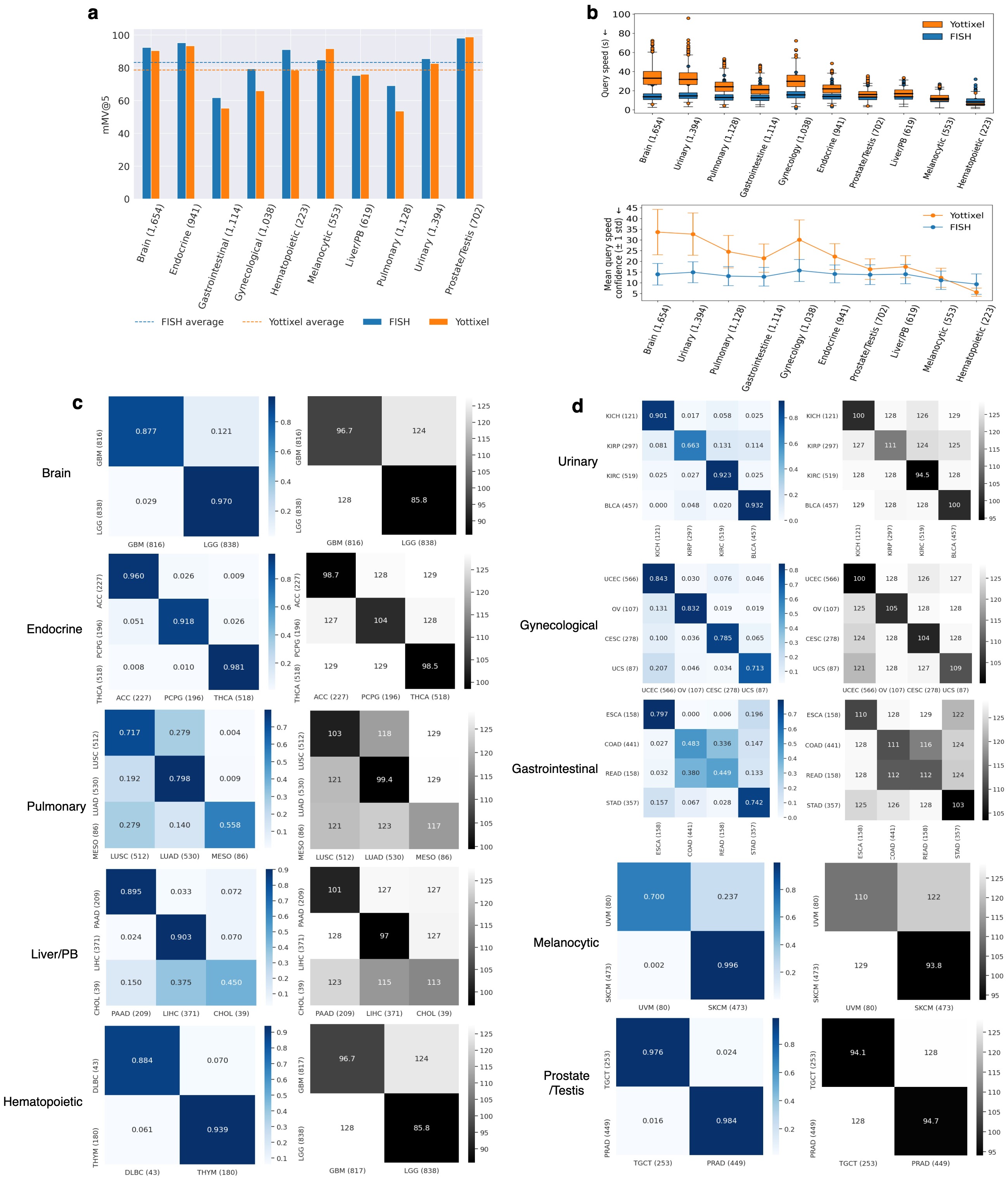}
    \caption{\textbf{Disease subtype retrieval in public cohorts.} \textbf{a} The marco-average mMV@5 of FISH and Yottixel on the TCGA anatomic sites. FISH has better performance in most sites. \textbf{b} Top: The query speed comparison between FISH and Yottixel for each site. Bottom: The mean confidence ($\pm1$ std) of query speed between FISH and Yottixel. It crucial to note that FISH is 2x more effective when the number of slide is over 1,000. Details study of speed was reported in \textbf{Speed and Interpretability}. \textbf{c-d} The confusion and hamming distance matrices provide insights into FISH's retrieval results. The x-axis and y-axis correspond to the ground truth and the model prediction respectively. The sharp and dark diagonal line in the confusion matrix and hamming distance matrix suggest that FISH can retrieve correct results in most of the cases. The PB in all figures stands for pancreaticobiliary and the number in the parentheses in \textbf{a-b} denote the number of slide in each site.}
    \label{fig_3}
\end{figure*}

\begin{figure*}[t!]
    \centering
    \includegraphics[width=0.85\textwidth]{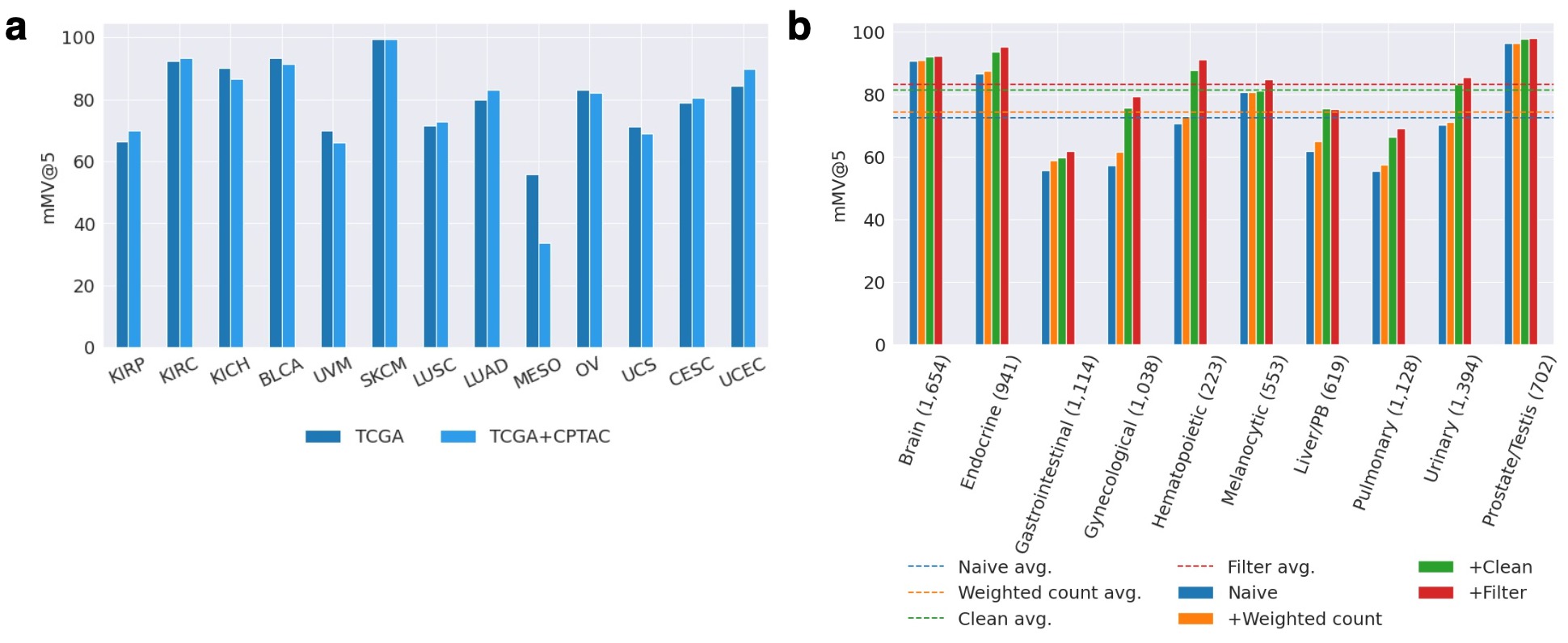}
    \caption{\textbf{Disease subtype retrieval in public cohorts and ablation study.} \textbf{a} The comparison between FISH on TCGA and TCGA + CPTAC cohorts. The performance does not vary before and after mixing with CPTAC cohorts for most of cases. The number of slides for each diagnosis on TCGA from left to right are: KIRP (297), KIRC (519), KICH (121), BLCA (457), UVM (80), SKCM (473), LUSC (512), LUAD (530), MESOS (86), OV (107), UCS (87), CESC (278) and UCEC (566). For TCGA + CPTAC cohort, the only differences are on KIRC (1,022), LUSC (1,191), LUAD (1,199), SKCM (756) and UCEC (1,110). \textbf{b} Ablation study result on the ranking module of FISH.  We observed that FISH has the best performance in the setting where all functions are applied ($+$Filter). The details of each setting was described in \textbf{Ablation Study} in our \textbf{Method}. PB denotes Pancreaticobiliary in liver/PB and the number in the parentheses denote the number of slide in each site.}
    \label{fig_4}
\end{figure*}

\vspace{-5mm}
\begin{spacing}{1.38}
\noindent\textbf{Disease subtypes retrieval in public cohorts}\\% \textbf{a} The average mMV@5 of FISH and Yottixle on all sites}\\
\noindent We reported the majority top-5 accuracy (mMV@5) for FISH and Yottixel as our main comparing metric and provided mAP@5 of FISH for reference. The mMV@5 evaluates how often the majority slide diagnosis in the top 5 results matches the query one, while mAP@5 measures how well the model can give slides with the same diagnosis as the query slide a higher rank in the retrieval results. We used mMV@5 as the primary metric for comparison as it is stricter than widely-used top-5 accuracy and also report mAP@5 because the results are considered correct only when the majority diagnosis in the retrieval agrees with the query. More details can be found in Online Methods.
We built the FISH pipeline on slides from each anatomic site and tested whether FISH can retrieve slides with correct diagnosis. Overall, we had better results than Yottixel (\textbf{Figure~\ref{fig_3}a}) in terms of macro-average mMV@5 within each site and across all sites. We believe macro-averaging is the appropriate measure here as the uncommon cases in an unbalanced real world histology database are as crucial as the common ones. For some sites such as Pulmonary, Gynecological, Urinary and Hematopoietic where the data distributions are skewed, FISH outperforms Yottixel in the uncommon diagnosis by large margin (47.68\% improvement on Pulmonary-MESO; 29\%, 16.3\%, 16.2\% improvement on Gynecological-UCS, Gynecological-CESC, Gynecological-OV, respectively; 14.2\% improvement on Urinary-KICH and 30.2\% improvement on Hematopoietic-DLBC). A detailed comparison is shown in \textbf{Table~\ref{tab:1}} and individual retrieval results are available in \textbf{Supplementary Table 1}. In addition, the speed advantage of FISH became pronounced especially after the number of slide in the database exceeded 1,000 (\textbf{Figure~\ref{fig_3}b}). The median query speed of FISH remains almost constant despite the growing number of slides, which is justified by our theoretical results. We perform more experiments to demonstrate that FISH is scalable to thousands of slides in a later section (\textbf{Speed and Interpretability}). Additionally,
% The ranking algorithm plays a crucial role in the success of FISH. Therefore we conduct an ablation study to validate the design of this module. We give a bird eye view of these modules and leave the detail in the Method. The ranking algorithm consists of three functions in order, which are a weighed uncertainty calculation (\textproc{Weighted-Uncertainty-Cal}), a result cleaning function (\textproc{Clean}), and a result filter (\textproc{Filter-By-Prediction}). The intuition behind this sequential process is that FISH gives higher priority to mosaic with highly certain retrieval results (achieved by \textproc{Weighted-Uncertainty-Cal}) and removes the ones with lower quality retrieval (achieved by \textproc{Clean}). To prevent FISH from overly relying on the most certain mosaic's retrieval results, we use top-K certain mosaic's prediction result as pseudo ground truth and remove the mosaic that disagrees with this pseudo label (achieved by \textproc{Filter-By-Prediction}). We show that FISH achieves the best performance by leveraging all function sequentially (red line in \textbf{Figure~\ref{fig_3}b}).
the ranking algorithm plays a crucial role in the success of FISH and applies three post-processing steps to the predictions. Therefore, we conducted an ablation study to validate these steps and showed that FISH achieved the best performance by including all the steps (red line in \textbf{Figure~\ref{fig_4}b}). The details of these steps are explained in the \textbf{Ablation study} in our Methods.

To further test the generalization ability of FISH, we combined several diseases (KIRC, UCES, SKCM, LUAD, and LUSC) in CPTAC with the TCGA data to test performance on a mixed public cohort with the results reported in \textbf{Figure~\ref{fig_4}a}. After combination, the distribution of the dataset in all sites became more skewed, but the performance of FISH did not vary substantially in most cases. This result further shows that FISH can address dataset unbalance commonly presented in the real world. The only exception was Pulmonary-MESO, for which the site where the disease is located was highly unbalanced. Individual retrieval results are available in \textbf{Supplementary Table 2}. Note that our VQ-VAE was only trained on TCGA data without observing slides in the CPTAC, which also showed the generalizability of our encoder.
\end{spacing}
\begin{spacing}{1.38}
We also created confusion matrices and hamming distance matrices in \textbf{Figure~\ref{fig_3}c-d} to gain more insight. The hamming distance calculates each diagnosis's average paired hamming distance in a given site, which helps further explain the trend behind the confusion matrix. More details about how we calculate these matrices are described in Method. By examining the confusion matrices, we can see a dark diagonal line, which suggested the majority results FISH retrieves match the queried diagnosis. The hamming distance matrices further explained the trends in the confusion matrices. The dark diagonal line shows the smallest hamming distance values in all sites, demonstrating that slides with different diagnosis are pushed far away in the hamming distance space. We can also use the matrices to explain why FISH performed worse in certain diseases. For example, query slide with diagnosis Liver-CHOL was more often confused with Liver-LIHC than Liver PAAD, which can be explained by the fact that the distance between Liver-CHOL and Liver-LIHC is smaller than to Liver-PAAD. Another example was the gastrointestinal site, where the diagonal values in the distance matrix were generally higher than in other sites, which explains why FISH performed worse in this site. We can apply similar logic to other sites and diseases.\\
\end{spacing}
\begin{table}[t!]
    \begin{minipage}{.5\linewidth}
    \centering
    \begin{tabular}{ccccc}
        \toprule
        \multirow{2}{*}[-1em]{Diagnosis} & \multirow{2}{*}[-1em]{\#slide} & \multicolumn{2}{c}{mMV@5} &  mAP@5\\
        \cmidrule{3-5}  \\
        \vspace{4mm}{} & & FISH & Yottixel  & FISH \\
        \midrule
        \textbf{Brain} \\
        GBM & 816 & 87.75 & \textbf{91.88}  & 88.31 \\
        LGG & 838 & \textbf{97.02} & 89.77  & 97.18 \\
        \midrule
        \textbf{Endocrine}\\
        ACC  & 227 &\textbf{96.04} & 93.83  & 95.96 \\
        PCPG  & 196 &\textbf{91.84} & 88.77  & 91.44 \\
        THCA  & 518 &\textbf{98.07} & 97.66  & 98.23 \\
        \midrule
        \textbf{Gastrointestinal}  \\
        COAD & 441 & 48.30 & \textbf{76.14}  & 58.07 \\
        ESCA  &158 &\textbf{79.75} & 59.87  & 79.83 \\
        READ & 158 & \textbf{44.94} & 10.19 & 48.75 \\
        STAD  & 357&\textbf{74.23} & \textbf{74.23}  & 77.09 \\
        \midrule
        \textbf{Gynecologic}\\
        UCEC  &566 &84.28 & \textbf{92.22}  & 86.74 \\
        CESC  &278 &\textbf{78.78} & 62.45  & 81.23 \\
        UCS  &87 &\textbf{71.26} & 42.22  & 72.51 \\
        OV  & 107&\textbf{83.18} & 66.98  & 83.88 \\
        \midrule
        \textbf{Hematopoietic}\\
        DLBC  &43 &\textbf{88.37} & 58.13  & 90.58\\
        THYM  &180 &93.89 &  \textbf{98.87}  & 95.99\\
        \bottomrule
    \end{tabular}
    \end{minipage}%
    \begin{minipage}{.4\linewidth}
    \centering
    \begin{tabular}{ccccc}
        \toprule
        \multirow{2}{*}[-1em]{Diagnosis} & \multirow{2}{*}[-1em]{\#slide} & \multicolumn{2}{c}{mMV@5} & mAP@5\\
        \cmidrule{3-5}\\
        \vspace{4mm}{} & &FISH & Yottixel  & FISH\\
        \midrule
        \textbf{Melanocytic}\\
        UVM  &80 & 70.00 & \textbf{83.75}  & 77.53\\
        SKCM  &473 & \textbf{99.58} & 99.57  & 99.69\\
        \midrule
        \textbf{Liver/PB}\\
        \textbf{}
        CHOL  &39 & \textbf{46.15} & 43.58  & 56.41 \\
        LIHC  &371 & 90.30 & \textbf{93.65}  & 91.16 \\
        PAAD  &209 &\textbf{89.47} & 91.04  & 89.41 \\
        \midrule
        \textbf{Pulmonary}\\
        LUAD  &530 &\textbf{79.81} & 70.96  & 82.26\\
        LUSC  &512 & 71.68 &  \textbf{81.70} & 73.52 \\
        MESO  &86 &\textbf{55.81} &  8.13 & 63.37 \\
              &    &       &                 &       \\
        \midrule
        \textbf{Urinary}\\
        BLCA  &457& 93.22 & \textbf{95.81}  & 93.38 \\
        KIRC  &519& \textbf{92.29} & 91.66  & 92.91 \\
        KICH  &121& \textbf{90.10} & 75.92  & 88.13 \\
        KIRP  &297 & 66.33 & \textbf{67.22}  & 68.62 \\
        \midrule
        \textbf{Prostate/Testis}\\
        TGCT  &254 & 97.64 & \textbf{99.21}  & 97.56 \\
        PRAD  &449 & \textbf{98.44} & 98.43  & 98.39 \\
        \bottomrule
    \end{tabular}
    \end{minipage}
    \caption{\textbf{Disease subtype retrieval on TCGA.} We compare FISH to Yottixel in terms of mMV@5 on the TCGA diagnostic whole slide images in 10 anatomic sites as this is the best result quoted from Yottixel's paper. FISH consistently performs better than Yottixel especially in the sites where the dataset is unbalanced (\textit{e.g.}, Gynecological and Pulmonary). We also report mAP@5 for FISH to evaluate whether FISH can give the desired slide higher rank. PB stands for pancreaticobiliary.}
    \label{tab:1}
\end{table}
\begin{figure*}[b!]
    \centering
    \includegraphics[width=\textwidth]{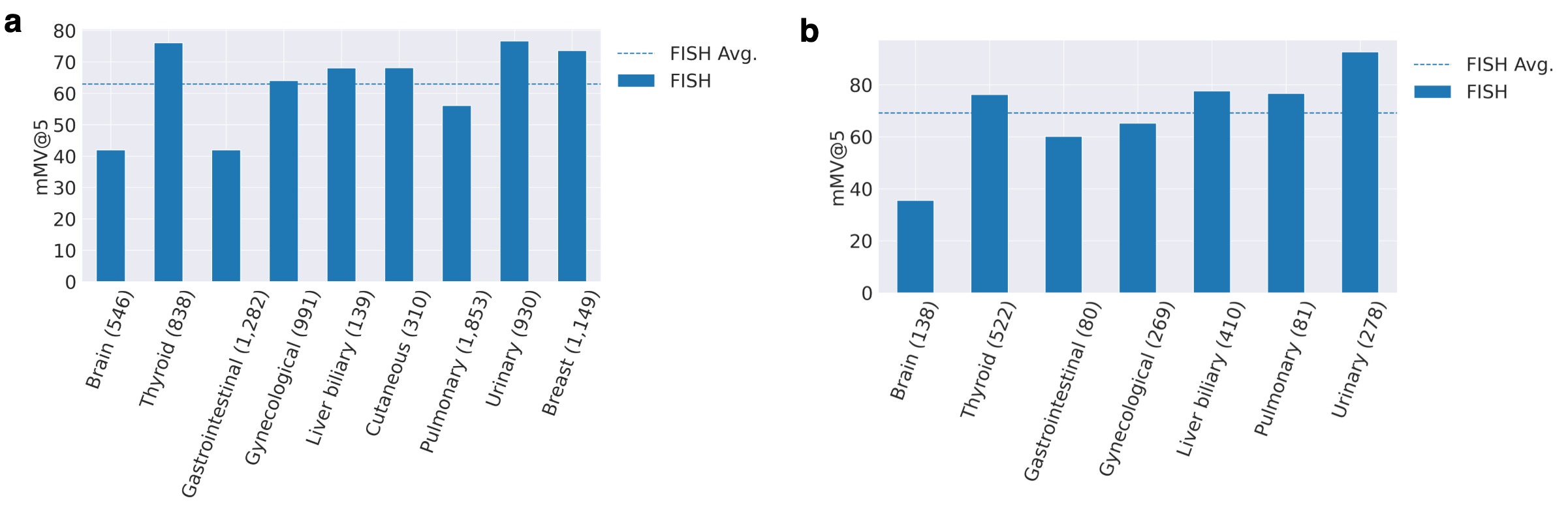}
    \caption{\textbf{Adapting to BWH independent test cohort.} \textbf{a} The average mMV@5 of FISH in each site in BWH general cohorts. \textbf{b} FISH's performance on rare type cancer in terms of average mMV@5 in each site. We observed that FISH has comparable performance in both general and rare disease cohort. The number in the parentheses denote the number of slide in each site.}
    \label{fig_5}
\end{figure*}
\begin{figure*}[t!]
    \centering
    \includegraphics[width=0.76\textwidth]{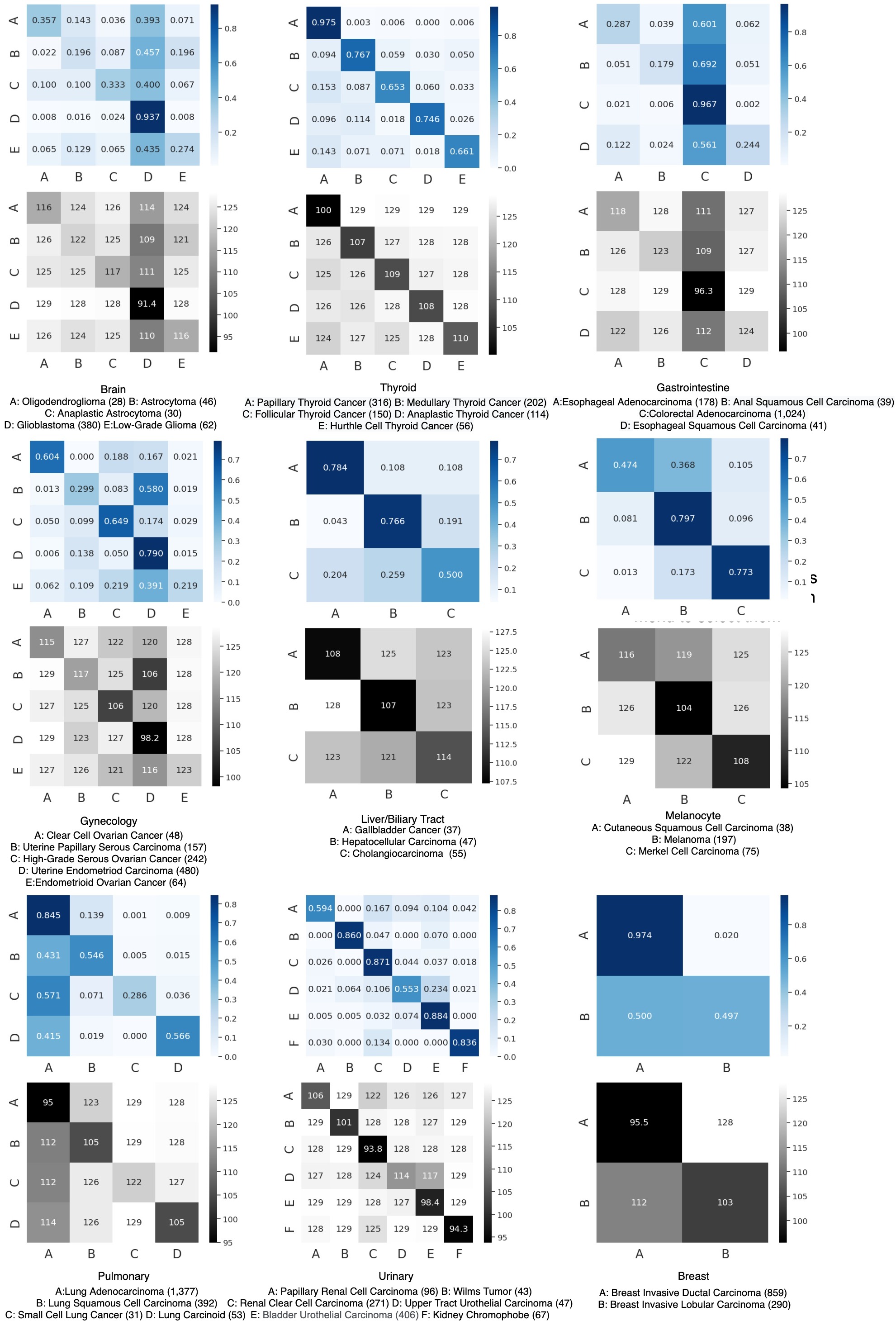}
    \caption{\textbf{Adapting FISH to independent BWH in-house whole slide images} The confusion matrix (blue color) and hamming distance matrix (grey color) of 37 disease from 9 sites in BWH in-house data. The x-axis and y-axis correspond to the ground truth and the model prediction respectively. The sharp and dark diagonal line in the confusion matrix and hamming distance matrix suggest that FISH can retrieve correct results in most of the cases. The alphabetical legends below each site name denote the diagnosis studied in the corresponding site.}
    \label{fig_6}
\end{figure*}
\begin{spacing}{1.37}
\vspace{-6mm}
\noindent\textbf{Adapting FISH to independent BWH in-house whole slide image.}\\
\noindent There are many variations in whole slide images (WSIs) across the institutions due to differences in the protocol of slide preparation and digitalization. Therefore, it is essential to validate our FISH trained on TCGA is robust to in-house data. We collected 8,035 diagnostic slides that contains 9 anatomic sites with 37 primary cancer subtypes from the WSI database at Brigham and Women’s Hospital. For each anatomic site, we built our pipeline separately and used mMV@5 as the main evaluation metric while provided mAP@5 for reference. FISH achieved an average $62.92\%$ mMV@5 across all sites (\textbf{Figure~\ref{fig_5}a}). It was especially successful in Urinary ($76.64\%$), Thyroid ($76.03\%$), Cutaneous ($68.13\%$), Liver/Biliary ($68.02\%$) and Gynecology($64.02\%$) as shown in \textbf{Figure~\ref{fig_6}} where the diagonal lines in both confusion and hamming distance matrix were relatively clear. We reported the detailed results in \textbf{Table~\ref{tab:3}} and individual retrieval results are available in \textbf{Supplementary Table 3}. Note that we did not fine-tune our encoder in this cohort, which shows the generalizability of encoder trained only on TCGA. To further investigate the clinical value of FISH, we conducted another experiment specifically for the rare type cancers by combing BWH cohorts and TCGA which results in 1,785 slides with 23 rare type cancers from 7 sites. FISH achieved $69.15\%$ in terms of mMV@5 (\textbf{Figure~\ref{fig_5}b}), which is comparable to the performance achieved on the general cohort in the previous experiment. The detailed results were described in \textbf{Table~\ref{tab:4}}. This is an encouraging result as it shows that if we create a whole slide database dedicated to rare disease, FISH can attain better performance. To the best of our knowledge, this is the first study that evaluates whole slide search engine on the rare diseases.
\end{spacing}

\begin{spacing}{1.37}
\vspace{-6mm}
\noindent\textbf{Rare disease retrieval}\\
\noindent The number of slides in rare diseases is usually less than the common ones, which makes the modern machine learning method challenging to train an efficient classifier upon it. The situation gets worse in some low-resource areas. To further investigate the clinical value of FISH, we conducted another experiment specifically for the rare type cancers by combing BWH cohorts and TCGA which results in 1,785 slides with 23 rare type cancers from 7 sites. FISH achieved $69.15\%$ in terms of mMV@5 (\textbf{Figure~\ref{fig_5}b}), which is comparable to the performance achieved on the general cohort in the previous experiment. The detailed results were described in \textbf{Table~\ref{tab:4}} and individual retrieval results are available in \textbf{Supplementary Table 4}. This is an encouraging result as it shows that if we create a whole slide database dedicated to rare disease, FISH can attain better performance. To the best of our knowledge, this is the first study that evaluates whole slide search engine on the rare diseases.
\end{spacing}

\begin{table}[t!]
    \centering
    \begin{tabular}{ccccc}
        \toprule
        Site  & Diagnosis & \#slide & mMV@5 & mAP@5\\
        %\cmidrule(lr){4-7}  \\
        \midrule
        \textbf{Brain} &
        Glioblastoma Multiforme & 380 & 93.68  & 92.34 \\
        & Low-Grade Glioma, NOS & 62 & 27.42   & 46.61 \\
        &Astrocytoma & 46 & 19.57  & 24.95 \\
        &Anaplastic Astrocytoma & 30 & 33.33 & 50.12\\
        &Oligodendroglioma & 28 & 35.71 & 50.99\\
        \midrule
        \textbf{Thyroid}&
        Papillary Thyroid Cancer & 316 & 97.47  & 97.42 \\
        &Medullary Thyroid Cancer & 202 & 76.73  & 80.26 \\
        &Follicular Thyroid Cancer & 150 & 65.33 & 69.57\\
        &Anaplastic Thyroid Cancer & 114 & 74.56  & 78.59  \\
        &Hurthle Cell Thyroid Cancer & 56 & 66.07  & 68.20 \\
        \midrule
        \textbf{Gastrointestinal}&
        Colorectal Adenocarcinoma & 1,024 & 96.68  & 95.67 \\
        &Esophageal Adenocarcinoma & 178 & 28.65   & 47.62  \\
        &Esophageal Squamous Cell Carcinoma & 41 & 24.39 & 26.54 \\
        & Anal Squamous Cell Carcinoma & 39 & 17.95  & 36.06\\
        \midrule
        \textbf{Gynecological}&
        Uterine Endometrioid Carcinoma & 480 & 78.96  & 82.35  \\
        &High-Grade Serous Ovarian Cancer & 242 & 64.88 & 64.71  \\
        &Uterine Papillary Serous Carcinoma & 157 & 29.94  & 44.16 \\
        &Endometrioid Ovarian Cancer & 64 & 21.88 & 35.70 \\
        &Clear Cell Ovarian Cancer & 48 & 60.42  & 58.86\\
        \midrule
        \textbf{Liver\&Biliary}&Cholangiocarcinoma & 55 & 49.09  & 57.23 \\
        & Hepatocellular carcinoma & 47 & 76.60  & 79.70 \\
        &Gallbladder Cancer & 37 & 78.38  & 73.77\\
        \midrule
        \textbf{Cutaneous}&
        Melanoma & 197 & 79.70  & 83.14  \\
        &Merkel Cell Carcinoma & 75 & 77.33  & 82.04\\
        &Cutaneous Squamous Cell Carcinoma & 38 & 47.37  & 59.22 \\
        \midrule
        \textbf{Pulmonary}&
        Lung adenocarcinoma & 1,377 & 84.46  & 85.91  \\
        &Lung squamous cell carcinoma & 392 & 54.59  & 60.96  \\
        &Lung Carcinoid & 53 & 56.60  & 74.55  \\
        &Small Lung Cell Cancer & 28 & 28.57  & 42.75  \\
        \midrule
        \textbf{Urinary}&
        Bladder Urothelial Carcinoma & 406 & 88.42 & 90.51 \\
        &Kidney renal clear cell carcinoma & 271 & 87.08  & 89.04  \\
        &Kidney renal papillary cell carcinoma & 96 & 59.38  & 62.27  \\
        &Kidney Chromophobe & 67 & 83.58  & 85.74  \\
        &Upper tract Urothelial Carcinoma & 47 & 55.32  & 61.73 \\
        &Wilms Tumor & 43 & 86.05  & 89.41  \\
        \midrule
        \textbf{Breast} &
        Breast Invasive Ductal Carcinoma & 859 & 97.44   & 98.29 \\
        & Breast Invasive Lobular Carcinoma & 290 & 49.66   & 50.55\\
        \bottomrule
    \end{tabular}
    \vspace{-2mm}
    \caption{\textbf{Adapting FISH to independent BWH in-house whole slide image.} The detail performance of FISH on 37 cancer types from BWH test cohort. The cancer in this table is a general cohort that contain both common and rare cancer subtype in each site.}
    \label{tab:3}
\end{table}

\begin{spacing}{1.37}
\vspace{-6mm}
\end{spacing}
\begin{table}[t!]
    \centering
    \begin{tabular}{cccccc}
        \toprule
        Site  & Diagnosis & \#slide & mMV@5 & mAP@5\\
        \midrule
        \textbf{Brain}& Astrocytoma & 46 & 32.61 & 46.12 \\
        & Anaplastic Astrocytoma  &30 & 40.00 & 50.79 \\
        & Oligodendroglioma & 28 &42.86 & 54.85 \\
        & Pilocytic Astrocytoma & 20 & 55.00& 64.48 \\
        & Anaplastic Oligodendroglioma & 14 & 7.14& 20.48\\
        \midrule
        \textbf{Thyroid}& Medullary Thyroid Cancer & 202 &80.20 & 83.21\\
        & Follicular Thyroid Cancer & 150 & 72.67& 78.86\\
        & Anaplastic Thyroid Cancer & 114 &77.19 &82.66 \\
        & Hurthle Cell Thyroid Cancer & 56 & 75.00&79.91\\
        \midrule
        \textbf{Gastrointestinal} & Esophageal Squamous Cell Carcinoma & 41 &53.66 & 61.59\\
        & Anal Squamous Cell Carcinoma & 39 &66.67 &69.47\\
        \midrule
        \textbf{Gynecological}& Uterine Papillary Serous Carcinoma & 157 & 93.63 &95.47\\
        & Endometrioid Ovarian Cancer & 64&43.75 &48.24\\
        & Clear Cell Ovarian Cancer & 48& 58.33&63.93\\
        \midrule
        \textbf{Liver Pancreaticobiliary}& Pancreatic Adenocarcinoma & 209 & 94.26& 95.06\\
        &Cholangiocarcinoma & 94 & 69.15&70.26\\
        &Pancreatic Neuroendocrine Tumor & 77 & 76.62&82.47\\
        &Gallbladder Cancer & 37 & 70.27&74.52\\
        \midrule
        \textbf{Pulmonary} & Lung Carcinoid & 53&96.23 &97.30 \\
        &Small Cell Lung Cancer & 28& 57.14&65.60\\
        \midrule
        \textbf{Urinary} & Kidney Chromophobe & 188 &97.34 & 97.61\\
        &Upper Tract Urothelial Carcinoma &47 &85.11 &85.53\\
        &Wilms Tumor &43 &95.35 &95.12\\
        \bottomrule
    \end{tabular}
    \caption{\textbf{FISH performance on  rare  type cancers.} The detail performance of FISH on 23 rare type cancers from BWH and TCGA cohort. We only use TCGA cohort for Cholangiocarcinoma, Pancreatic Adenocarcinoma and Kidney Chromophobe.}
    \label{tab:4}
\end{table}

\begin{figure*}[t!]
    \centering
    \includegraphics[width=0.9\textwidth]{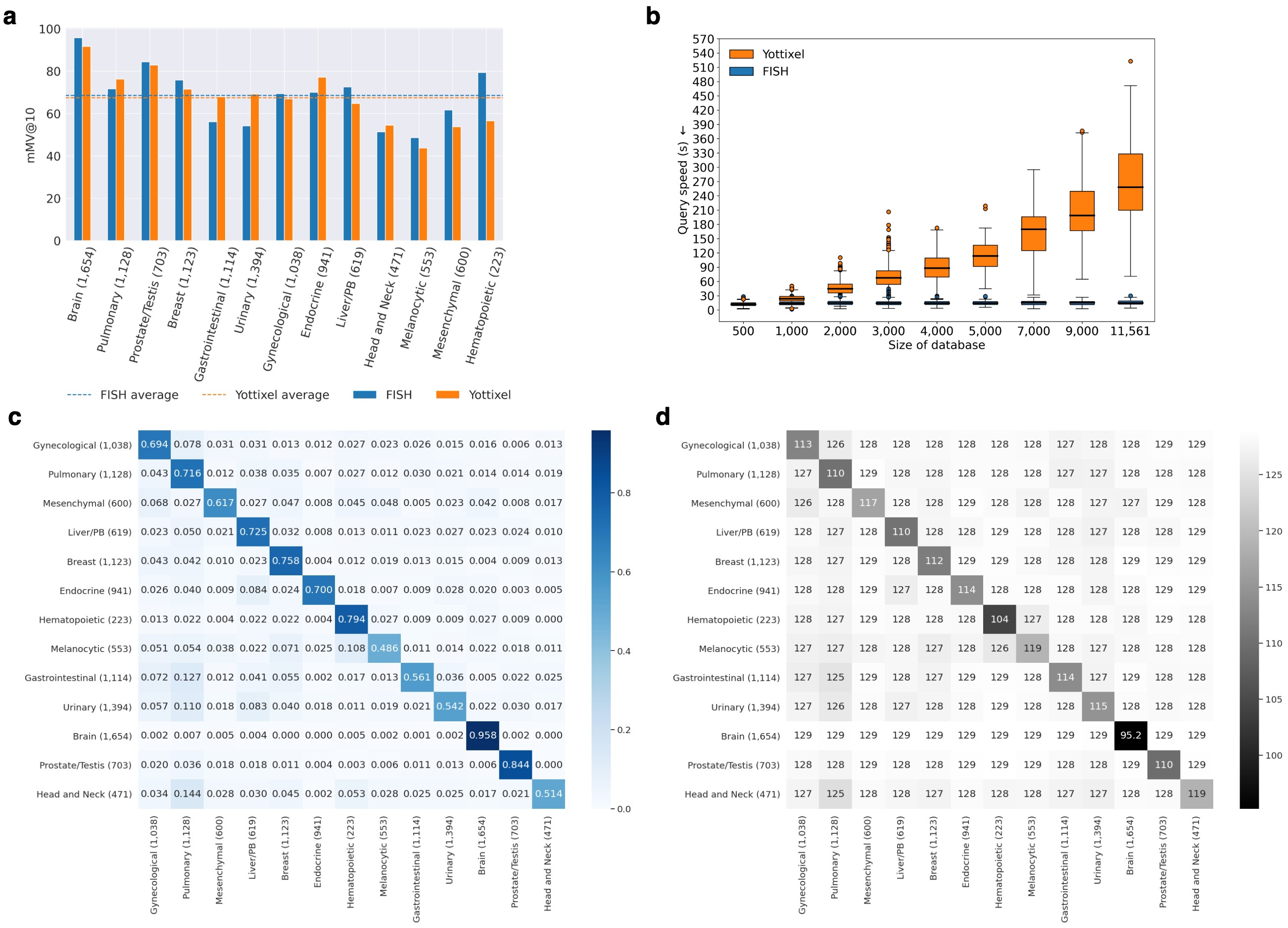}
    \caption{\textbf{Performance on anatomic site retrieval and speed.} \textbf{a} The mMV@10 comparison between FISH and Yottixel. \textbf{b} The speed comparison between FISH and Yottixel in the TCGA anatomic site retrieval. FISH is faster than Yottixel by 15x when the number of slide is over 10,000. Please refer to \textbf{Speed and Interpretability} fore more details. \textbf{c, d} The confusion matrix and hamming distance matrix of FISH on anatomic site retrieval. The x-axis and y-axis correspond to the ground truth and the model prediction respectively. The sharp diagonal line in both matrix showed that FISH can retrieval the correct results and push the dissimilar ones in most cases. The number in the parentheses denotes the number of slide in each site. PB stands for pancreaticobiliary.}
    \label{fig:organ_speed}
\end{figure*}

\begin{spacing}{1.4}
\noindent\textbf{Anatomic sites retrieval} \\
Although we usually know the anatomic site where the tissue is resected nowadays, it is still possible that the site information is missed in some old whole slide image database. A search engine that can return the slide with the same sites is beneficial to archive the database. We used the diagnostic slide from TCGA and follow the paper\cite{kalra2020yottixel} to group slides into 13 categories which results in 11,561 whole slide images. We built FISH pipeline on this database and the goal was to retrieve slides with the same anatomic site as the queried one. FISH achieved $68.52\%$ mMV@10 on average which was slightly better than Yottixel ($67.42\%$) (\textbf{Figure~\ref{fig:organ_speed}a}). We compared the mMV@10 in this experiment as this was the best performance reported in the Yottixel's paper\cite{kalra2020yottixel}. It important to note that FISH is over 15$\times$  faster than Yottixel as shown in the rightmost box plot in the \textbf{Figure~\ref{fig:organ_speed}b}) although the gap between two methods is small. The detail study of speed between FISH and Yottixel can be found in \textbf{Speed and Interpretability} and individual retrieval results are available in \textbf{Supplementary Table 5}.
\end{spacing}
\vspace{-5mm}
\begin{figure*}[t!]
    \centering
    \includegraphics[width=0.85\textwidth]{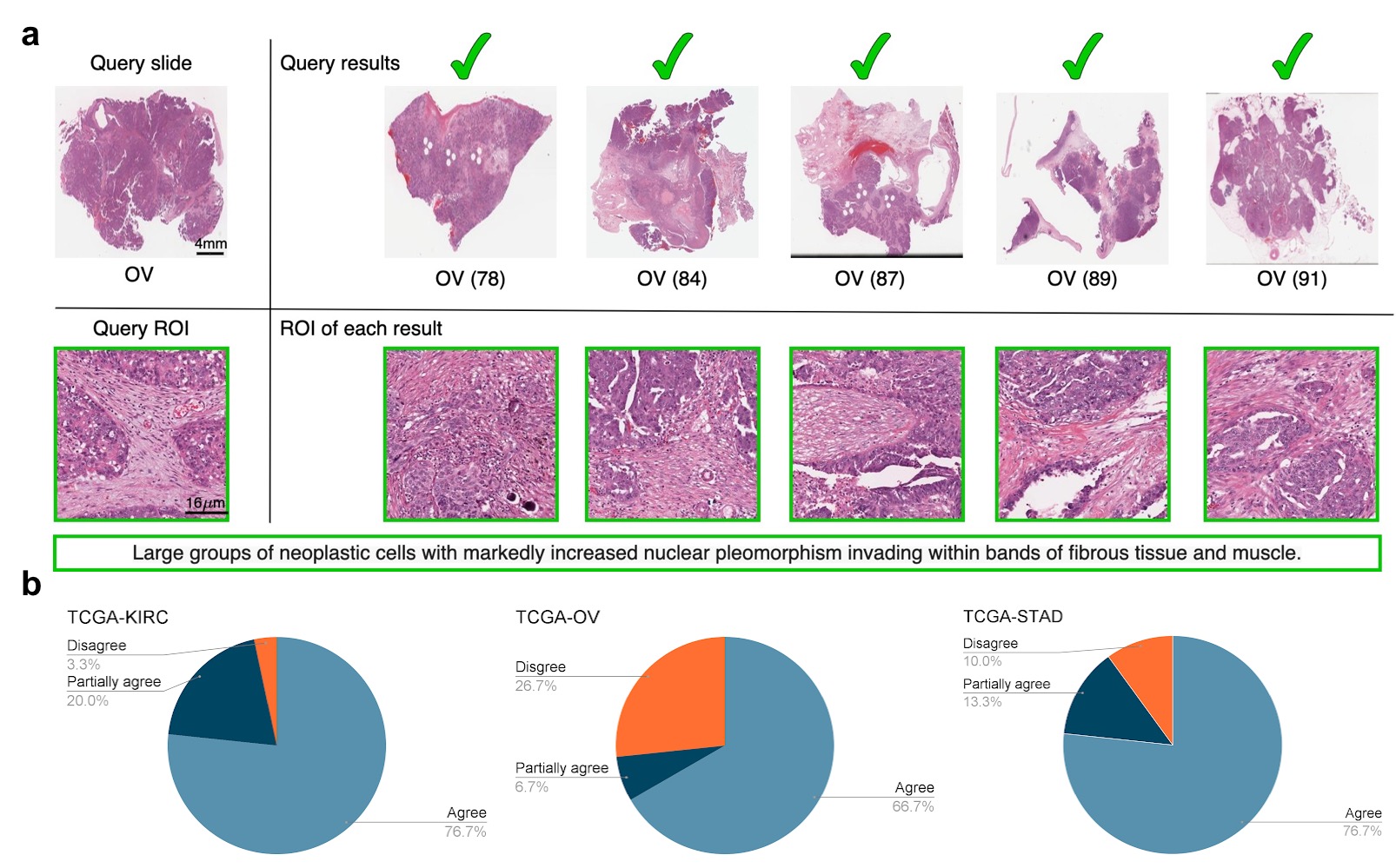}
    \caption{\textbf{Interpretability and speed.} \textbf{a} FISH outputs the region of interest that is useful to define the similarity of a cancer type in both query slide and each slide in the results. The number in parentheses is the hamming distance between the query slide and each result, determined by the identified ROI in each WSI. More examples can be found in \textbf{Extended Data Figure 1-4}. \textbf{b} We studied KIRC, OV and STAD in TCGA cohort with $92.29\%$, $83.18\%$ and $74.23\%$ mMV@5 score. Each study contains 30 randomly selected queries and each query contains 1-5 ROIs. The pathologist rates agree, partially agree and disagree based on whether the regions contain tumors. The ratio of agree plus partially agree is over $70\%$ in all studies.}
    \label{fig_7}
\end{figure*}
\begin{spacing}{1.4}

\noindent\textbf{Analysis of Speed and interpretability} \\
Speed and interpretability are essential properties of consideration for whole slide image search engines in addition to accuracy. Fast search speeds enable the usability of search engine in large databases in the digital pathology era and interpretability makes the system easy to debug and more robust to unexpected errors. We demonstrated that FISH has all these desired properties in this section.

We show how FISH interpret the results of query slide in \textbf{Figure~\ref{fig_7}a}. For a query slide, FISH returns the regions in the slide that are useful for defining the similarity of cancer type. This is allows us to examine these regions and ensure the search system returns the results based on some evidence agreed by pathologist instead of unmeaningful regions such as debris. More examples are shown in the \textbf{Extended Data Figure~1-4} We conducted three interpretation studies using TCGA-KIRC, TCGA-OV and TCGA-STAD respectively to understand FISH's interpretability across different levels of performance (in terms of differences in mMV@5 scores). For each study, we randomly selected 30 queries which contained at least 1 correct retrieval in the results and then extracted the ROIs found in the query slide. We asked a pathologist to rate whether the ROIs agree with their judgement by ``agree", ``partially agree" (\textit{i.e.}, if the pathologist agrees with at least one of ROIs) and ``disagree". For example, in the study of TCGA-KIRC, the prompt was whether the ROI contain features of KIRC. The results were shown in \textbf{Figure~\ref{fig_7}b}. The key finding was that the ratio of agree plus partially agree exceeded $70\%$ in all studies.

We used the same TCGA data in the anatomic site retrieval experiment to evaluate query speed. We applied weighted sampling to to select slides from each site to create database of size 500, 1,000, 2,000, 3,000, 4,000, and 5,000, 7,000, 9,000 together with the original dataset of size 11,561. We implemented both method in Python and evaluated them on the same machine for a fair comparison.  The average query speeds of both methods are reported in \textbf{Figure~\ref{fig:organ_speed}b}. Since we observed that Yottixel was inefficient beyond 3,000 slides, we use the same 100 queries sampled from the databases to calculate the average query speed of FISH and Yottixel instead of using all data when the size of the database goes beyond 3,000. On the contrary, the average query speed of FISH remained almost constant with low variances throughout the experiments, which agree with our theoretical results. This result is highly encouraging as it demonstrates that FISH can scale with the growing number of slides in the digital pathology era while maintaining a relatively constant query speed.
\end{spacing}
\vspace{-5mm}
% \begin{figure*}[t!]
%     \centering
%     \begin{minipage}{.4\textwidth}
%     \centering
%     \includegraphics[width=\textwidth]{figs/FISH_speed.jpg}
%     \caption{\textbf{Speed evaluation.} Average query speed comparison between FISH and Yottixel. The average query time of FISH grows almost $O(1)$ which matches the theoretical result. The fluctuation happens because it is possible that the number of mosaic per slide in certain anatomic site with fewer number of slide is larger than that in the others.}
%     \label{fig_7}
%     \end{minipage}%
%     \hspace{5mm}
%     \begin{minipage}{.45\textwidth}
%         \centering
%         \includegraphics[width=\textwidth]{figs/FISH_TCGA_ablatoin.jpg}
%     \vspace{-9mm}
%     \caption{\textbf{Ablation study of FISH on TCGA.} We evaluate the ranking module of FISH in TCGA cohort by mMV@5. The results that includding all components (red bar) achieve the best performance in most of cases.}
%     \label{fig_8}
%     \end{minipage}
% \end{figure*}
\begin{figure*}[t!]
    \centering
    \includegraphics[width=0.75\textwidth]{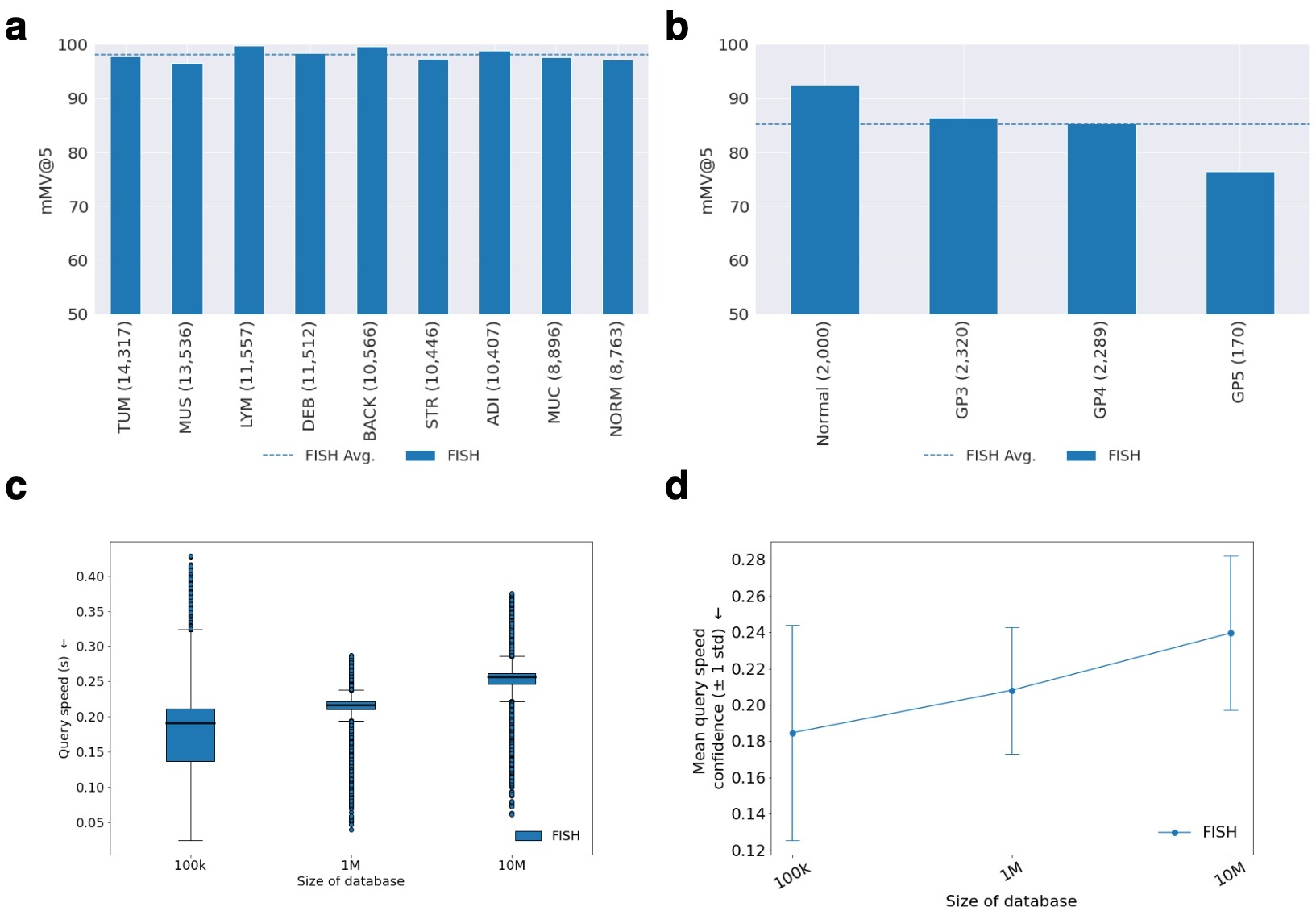}
    \caption{\textbf{Patch level retrieval.} \textbf{a} FISH's mMV@5 score on each type of tissue in Kather100k dataset. \textbf{b} FISH's mMV@5 score on prostate tissue with different Gleason score in the in-house prostate data. \textbf{c, d} FISH query speed and its mean confidence on the real kather100k data and two augmented ones (1M and 10M). Note the low variance in the kather1M and kather10M is due to the way we do data augmentation. An outlier with low query speed in original kather100k will get 10 and 100 neighbors after augmentation in kather1M and kather10M. The examples of retrieval can be found in \textbf{Extended Data Figures~5-6}.}
    \label{fig_8}
\end{figure*}
\begin{spacing}{1.5}
\newpage
\noindent\textbf{Patch level retrieval} \\
We show that FISH can also perform patch-level retrieval in O(1) query speed, although it is not designed for the task. In this task, we view each patch query as a single mosaic fed into the FISH search pipeline. Since there is only one mosaic, there is no need for the ranking module. We get the top-K results by directly sorting the predictions by their hamming distance. We evaluated FISH on Kather100k without color normalization (NCT-CRC-HE-100K-NONORM)\cite{kather2019predicting} and BWH in-house prostate data.

Kather100k contains 9 types of tissue of size $224 \times 224$ and in-house prostate data contain 4 types of annotations (Normal tissue, Gleason score 3, Gleason score 4 and Gleason score 5) of size $256\times256$ cropped from slides at 20$\times$. We resized them to $1024 \time 1024$ before feeding into our pipeline. We observed that FISH achieves $98.09\%$ (\textbf{Figure~\ref{fig_8}a}) and $85.15\%$ (\textbf{Figure~\ref{fig_8}b}) macro-average mMV@5 on Kather100k and in-house prostate data. The individual retrieval results are available in \textbf{Supplementary Table 6} and \textbf{Supplementary Table 7} for Kather100k and in-house prostate. More example results can be found in \textbf{Extended Data Figures~5-6}. We also conducted the speed test on our Kather100k data. To efficiently curate larger dataset for testing speed, we applied data augmentation by directly adding noise to the latent code of each patch from VQ-VAE encoder instead of raw image data. For each latent code, we added a binary array whose element equals to $0$ and $1$ with probability $0.95$ and $0.05$ respectively as noise to augment the data. All augmented data shared the same texture feature with the original one. We curated the dataset kather1M and kather10M by augmenting each patch 10 and 100 times respectively and used 100k patch in original data as query to test the query speed. We observed that the median query speed of FISH ranges from 0.15 to 0.25s and remains unaffected all the way to 10M. \textbf{Figure~\ref{fig_8}c-d}. Note that the work most related to our study reports 25s per query on the 10M patch data\cite{hegde47822}.
\end{spacing}

\begin{spacing}{1.5}
\vspace{-9mm}
\section*{\large{Discussion}}
\vspace{-6mm}
In summary, we show that FISH addresses several key challenges in whole slide image search: speed, accuracy, and scalability. Our experiments demonstrate that FISH is an interpretable histology image search pipeline, achieving $O(1)$ constant speed search after training with only slide-level labels. This constant search speed and lack of reliance on pixel-level annotations will only become increasingly important as institutions' WSI repositories grow to hundreds of thousands or millions of slides. We also showed that FISH has strong performance on unbalanced datasets commonly seen in real world histopathology, and can generalize to independent test cohorts, rare diseases, and can even be used as a search engine for patch retrieval.

To the best of our knowledge, our study presents the first search pipeline that is evaluated on the most diverse and largest dataset of diagnostic slides to date, while also reporting speed performance, an essential metric for a histology search engine\cite{komura2018machine}. We are also the first to evaluate the whole slide image search engine on rare type cancers. Additionally, FISH is the first search pipeline that can provide interpretable results for interrogation by pathologists.

Although our combination of a VQ-VAE and the vEB tree is key to success in our method, this approach is limited by the expressiveness of the integer indices created in this way. The accuracy of the method could be increased by increasing the length of these indices, but at the cost of increasing the size of metadata needed for searching and decreasing the speed of the search itself, as the vEB tree would need to visit more neighbors before finding the optimal candidates to return. One line of future work is to design a better indexing system whereby the distances would be more semantically relevant to expedite the searching process of the vEB tree, though one can imagine a system where institutions tweak the index for more speed or greater accuracy, depending on their needs. In addition, due to the limited access to large annotated patch data, the performance of FISH on large scale patch level retrieval has not yet been fully investigated. As such, evaluating FISH on millions or even billions annotated patch data is also a promising future direction.

Human-in-the-loop computing has been identified as a potential way to bring deep learning-based applications for medical images closer to the clinic\cite{budd2021survey}. Allowing end-users to give feedback and then using that feedback to iteratively refine the system can allow algorithms to better generalize to unseen data\cite{budd2021survey}. Many deep learning-based medical image segmentation models have utilized this concept\cite{wang2018interactive, wang2018deepigeos,amrehn2017ui}, but it is not commonly used in histology image search systems. In our study, we have shown that FISH can return interpretable semantic descriptors for both query and result slides, making it feasible to build a feedback loop into FISH whereby pathologists could agree or disagree with semantic descriptors to refine or expand the search without any additional training or fine-tuning. This may be especially useful in complex settings such as rare disease retrieval where finding additional data to improve search results may be impossible. By providing researchers and pathologists with a novel and efficient way of searching, sharing, and accessing knowledge and by leveraging human-in-the-loop computing, FISH shows promise in the seamless integration into the digital pathology workflow and demonstrates its potential role in medical education, research, and even the clinical setting.

\end{spacing}
\vspace{-4mm}
\section*{\large{Online Methods}}
\begin{spacing}{1.3}
\vspace{-4mm}
\noindent\textbf{FISH} \\
FISH is a histology image search pipeline that addresses the scalability issue of speed, storage, and pixel-wise label scarcity. It builds upon a set of mosaics preprocessed from whole slide images without pixel-wise labels to save storage and labelling cost and achieves $O(1)$ search speed by leveraging the benefits from discrete latent code of VQ-VAE, Guided Search Algorithm, and Ranking Algorithm. We present these essential components of FISH in this section.
\vspace{-4mm}

\noindent\textit{Discrete Latent Code of VQ-VAE.} VQ-VAE\cite{oord2017neural} is a variant of VAE that introduces a training objective that allows discrete latent code. Let $\boldsymbol{e}\in \mathbb{R}^{K\times D}$ be the latent space (\textit{i.e.}, codebook) where $K$ is the number of discrete codeword and $D$ is the dimension of codeword. We use $K=128$ and $D=256$ in our experiment. To decide the codeword of the given input, an encoder $q$ encodes input $\boldsymbol{x}$ as $\boldsymbol{z_e}(\boldsymbol{x})$. The final codeword $\boldsymbol{z}_{q}(\boldsymbol{x})$ of $\boldsymbol{x}$ and the training objective function are given by

\vspace{-12mm}
\begin{align}
    \boldsymbol{z}_{q}(x) = \boldsymbol{e}_{k},\textrm{ where } k = \textrm{argmin}_{j}\left\lVert\boldsymbol{z_e}(\boldsymbol{x}) - \boldsymbol{e}_{j}\right\rVert,
\end{align}
\begin{align}
    \log p(\boldsymbol{x}|\boldsymbol{z}_{q}(\boldsymbol{x})) + \left\lVert \textrm{sg}[\boldsymbol{z_e}(\boldsymbol{x})] - \boldsymbol{e}\right\rVert + \beta\left\lVert \boldsymbol{z_e}(\boldsymbol{x}) - \textrm{sg}[\boldsymbol{e}]\right\rVert,
\end{align}
where $\beta$ is a hyperparamter and sg denotes stop gradient operation.  A stop gradient operation acts as identify function in forward pass while have zero gradient during the backward. The first term in the objective function optimizes the encoder and decoder to have a good reconstruction, the second term is used to update the codebook, and the third term is used to prevent the encoder's output from diverting the latent space too far. The detail architecture of our VQ-VAE is shown in \textbf{Extended Data Figure~7}. We reordered the codebook based on the value of the first principle component and change the latent code accordingly as we found the reordered codebook can provide more semantic concept of the original input image (\textbf{Extended Data Figure~8}).
\end{spacing}

\vspace{-8mm}

\begin{spacing}{1.5}
\noindent\textit{Feature Extraction, Index Generation and Index Encoding.}
We show how each mosaic $i$ can be represent by a tuple $(m_{i}, h_{i})$ composed of mosaic index $m_{i}$ and mosaic texture feature $h_{i}$.  To get $m_{i}$, we encode and re-map the latent code $\boldsymbol{z}_{i}$ by the encoder and reorderd codeook in VQ-VAE. The index $m_{i}$ is determined by following equations.

\vspace{-16mm}
\begin{align}
    \boldsymbol{z}_{i,1} &= \textproc{AvgPool(2,2)}(\boldsymbol{z}_{i})\\
    \boldsymbol{z}_{i,2} &= \textproc{AvgPool(2,2)}(\boldsymbol{z}_{i,1})\\
    \boldsymbol{z}_{i,3} &= \textproc{AvgPool(2,2)}(\boldsymbol{z}_{i,2})\\
    m_{i} &= \textproc{Sum}(\boldsymbol{z}_{i,1}) + \textproc{Sum}(\boldsymbol{z}_{i,2})\times 10^{6} + \textproc{Sum}(\boldsymbol{z}_{i,3})\times 10^{11}\\
    \textproc{Sum}(z_{i,1}) &\in [0, 130048], \textproc{Sum}(z_{i,2}) \in [0,32512], \textproc{Sum}(z_{i,3}) \in [0,8128]
\end{align}
We insert each $m_i$ into vEB tree for fast search. Note that the time complexity of all operations in vEB tree are $O(\log\log(M))$. Based on the property of vEB tree, $M$ can be determined by

\vspace{-12mm}
\begin{align}
    M = 2^{x} > \textrm{max}(m_{i})~,
\end{align}
where $x$ is the minimum integer that makes inequality hold. Since our codebook size ranges from 0 to 127, we can determine the maximum summation $\textproc{Sum}(z)$ in each level. Solving the inequality, we can get the minimum of $M$ to satisfy the inequality is $M = 1125899906842624$. Because $M$ is a constant that only depends on the index generation pipeline, our search performance is $O(1)$. To get $h_{i}$, we use DenseNet128 to extract feature from $1024\times1024$ patch at 20$\times$ then follow the algorithm proposed in the paper\cite{kalra2020yottixel} to binarize it.
\vspace{-4mm}

In addition to creating the tuple to represent the mosaic, we also make a hash table $H$ with $m_{i}$ as key and value as metadata of the mosaic. The metadata includes the texture feature $h_{i}$, the slide name associated with the mosaic, the coordinate of the slide where this mosaic is cropped, the slide file format, and the diagnosis of the slide. Note that different mosaics could share the same key. In this case, the value is a list that stores all metadata.

\noindent\textit{Guided Search Algorithm.} Given the query slide $I$ represented as $I = \{(m_{1},h_{1}),(m_{2}, h_{2}),\dots,(m_{k}, h_{k})\}$ with $k$ mosaics where each tuple is composed of the index of mosaic $m_{i}$ and its texture feature $h_{i}$, we apply \textproc{Guided-Search} to each tuple and return the corresponding results $r_{i}$. The output takes the form of $R_{I} = \{r_{1}, r_{2},\dots, r_{k}\}$. Each $r_{i} =\{(p_{i1}, \mu_{i1}), (p_{i2}, \mu_{i2}),\dots,(p_{in}, \mu_{in})\}$ is a set of tuple consists of the indices of the similar mosaics $(p_{i1}, p_{i2},\dots, p_{in})$ in the database and the associated information $(\mu_{i1},\mu_{i2},\dots,\mu_{in})$. Each element $\mu_{il}$ contains the hamming distance between $p_{il}$ and the query mosaic along with metadata of $p_{il}$, including the diagnosis and site of slide where $p_{il}$ from, the $(x,y)$ position where $p_{il}$ is cropped, and the slide file format.
\vspace{-4mm}

The drawback to use only $m_{i}$ for query is that the current mosaic index is sensitive to the minor change in $\boldsymbol{z}_{i,3}$. For example, a mosaic that differs from another by $1$ incur $10^{11}$ difference, which put two mosaics far away to be searched by the vEB tree. To address this issue, we create a set of candidates indices $m_{i,c+}$ and $m_{i,c-}$ along with the original $m_{i}$ by adding and subtracting an integer $C$ for $T$ times from $\textproc{Sum}(\boldsymbol{z}_{i,3})$. We call helper functions \textproc{Forward-Search} and \textproc{Backward-Search} to search the neighbor indices in $m_{i,c+}$ and $m_{i,c-}$ respectively. Both functions are only include the neighbor indices whose hamming distance from the query $h_{i}$ is smaller than a threshold $\theta_{h}$. The details of algorithms are shown from \textbf{Algorithm 1-3}.
\begin{singlespace}
\begin{algorithm}
    \begin{algorithmic}
        \State $H \gets hash\ table$ \Comment{hash table with key as mosaic index and value as metadata}
        \State $C, T \gets 50\times 10^{11}, 10$ \Comment{Integer and number of times for addition and deduction}
        \State $\theta_{h} \gets 128$ \Comment{Threshold of the hamming distance between query mosaic index and the neighbor}
        \State $k_{succ}, k_{pred} \gets 375$ \Comment{Number of time to call \textit{vEB.Successor() and vEB.Predecessor()}}
        \Function{Guided-Search}{$m_{i}, h_{i}, C, T, \theta_{h}, k_{\textrm{pred}}, k_{\textrm{succ}}, H, vEB$}
            \State $m_{i,c+}, m_{i,c-}, results \gets \{\}, \{\}, \{\}$
            \State $V \gets \{\}$
            \State $m_{i,c+}.insert(m_{i})$
            \For{$ t \gets 1,2,...\thinspace, T$}
                \State $m_{tmp+}, m_{tmp-} \gets m_{i} + t \times C, m_{i} - t \times C$
                \State $m_{i,c+}.insert(m_{tmp+})$
                \State $m_{i,c-}.insert(m_{tmp-})$
            \EndFor
            \State $results_{+}, V \gets $\textproc{Forward-Search}($m_{i,c+}, k_{succ}, \theta_{h},V,H,vEB$)
            \State $results_{-} \gets  $\textproc{Backward-Search}($m_{i,c-}, k_{pred}, \theta_{h},V,H,vEB$)
            \State $results.insert(results_{+})$
            \State $results.insert(results_{-})$
        \State $results \gets $ \textproc{Sort-Ascending}($results, key=results.hammingdistance$)
        
    \State \Return{$results$}
    \EndFunction
\end{algorithmic}
\caption{Guided Search Algorithm}
\label{algo1}
\end{algorithm}
\end{singlespace}
\begin{singlespace}
\begin{algorithm}[t!] 
    \begin{algorithmic}
        \Function{Forward-Search}{$m_{i,c+}, k_{succ},\theta_{h},V,H,vEB$}
            \State $res_{+} \gets \{\}$
            \For{$i_{+}$ in $m_{i,c+}$}
            \State $succ\_cnt, succ_{prev} \gets 0, i_{+}$
            \While{$succ\_cnt < k_{succ}$}
            \State $succ \gets vEB.Successor(succ_{prev})$
            \If {$succ \in V$ or $succ$ is empty}
                \State break
            \ElsIf{$H[succ].len() == 0$}
\State// The case when the patient is identical to query slide $I$
                \State $succ_{prev} == succ$
                \State continue
            \Else
                \State // Find the mosaic with smallest hamming distance in the same key
                \State $dist_j,j \gets \textproc{argmin}_{j}(\textproc{Hamming-Distance}(h_{i},H[succ])$
            \EndIf
            \If{$dist_j < \theta_{h}$}
                \State $V.insert(succ)$
                \State $meta \gets H[succ][j]$
                \State $res_{+}.insert((dist_j, meta))$
            \EndIf
            \State $succ\_cnt, succ_{prev} \gets succ\_cnt + 1, succ$
            \EndWhile
            \EndFor
        \Return{$res_{+}, V$}
    \EndFunction
\end{algorithmic}
\caption{Foward Search Algorithm}
\label{algo2}
\end{algorithm}
\end{singlespace}
\begin{singlespace}
\begin{algorithm}[h!] 
    \begin{algorithmic}
    \Function{Backward-Search}{$m_{i,c-},k_{pred},\theta_{h},V,H,vEB$}
        \State $res_{-} \gets \{\}$
        \For{$i_{-}$ in $m_{i,c-}$ }
        \State $pred\_cnt, pred_{prev} \gets 0, i_{-}$
        \While{$pred\_cnt < k_{pred}$}
        \State $pred \gets vEB.Predecessor(pred_{prev})$
        \If {$pred \in V$ or $pred$ is empty}
            \State break
        \ElsIf{$H[pred].len() == 0$}
            \State // The case when the patient is identical to query slide $I$
            \State $pred_{prev} == pred$
            \State continue
        \Else
            \State // Find the mosaic with smallest hamming distance in the same key
            \State $dist_j,j \gets \textproc{argmin}_{j}(\textproc{Hamming-Distance}(h_{i},H[pred]))$
        \EndIf
        \If{$dist_j < \theta_{h}$}
            \State $V.insert(pred)$
            \State $meta \gets H[pred][j]$
            \State $res_{-}.insert((dist_j, meta))$
        \EndIf
        \State $pred\_cnt, pred_{prev} \gets pred\_cnt + 1, pred$
        \EndWhile
        \EndFor
    \Return{$res_{-}$}
    \EndFunction
\end{algorithmic}
\caption{Backward Search Algorithm}
\label{algo3}
\end{algorithm}
\end{singlespace}
\vspace{-10mm}
\noindent\textit{Results Ranking Algorithm.} Our ranking function \textproc{Ranking} (\textbf{Algorithm~\ref{algo4}}) takes the results $R_{I}=\{r_{1}, r_{2},\dots,r_{k}\}$ from \textproc{Guided-Search} as the input. The output is the top 5 similar slides given the query slide $I$. The intuition of \textproc{Ranking} is to find the most promising mosaics in $R_{I}$ based on the uncertainty. It relies on three helper functions, which are \textproc{Uncertainty-Cal} (\textbf{Algorithm~\ref{algo5}}), \textproc{Clean} (\textbf{Algorithm~\ref{algo6}}) and \textproc{Filtered-By-Prediction} (\textbf{Algorithm~\ref{algo7}}).
\begin{singlespace}
\begin{algorithm}[t!]
    \caption{Results Ranking Algorithm} \label{algo4}
    \begin{algorithmic}
        \Function{Ranking}{$R_{s}$, $D\_inv$, $N$, $K$}
        \If{$R_{s}$ is empty}
            \Return{}
        \EndIf
        \State $D\_inv \gets \textproc{Normalize}(D\_inv, N)$\Comment{Normalize the reciprocal of diagnosis count so that the sum is equals N. $N=10$ for the fixed site experiment and $N=30$ for the anatomic site one.}
        \State $S_{lb}, S_{m} \gets \{\},\{\}$
        \State $S_{l} \gets \{\}$
        \For{each mosaic's results $r_{i}$ in $R_{S}$}
            \If{$r_{i}$ is not empty}
                \State $Ent, label\_cnt, dist \gets $\textproc{Uncertainty-Cal}($r_{i}, D\_inv$)
                \State $S_{lb}.insert(i, label\_cnt)$
                \State $S_{m}.insert((i, Ent,dist, r_{i}.len()))$
                \State $S_{l}.insert(r_{i}.len())$
            \Else
                \State continue
            \EndIf
        \EndFor
        \State $S_{m} \gets \textproc{Clean}(S_{m}, S_{l}$)
        \State $f \gets \textproc{Filtered\_By\_Prediction}(S_{m}, S_{lb}$)
        \State $R_{ret}, V \gets \{\}, \{\}$
        \For{$e$ in $S_{m}$}
            \State $uncertainty, i \gets e.Ent, e.i$
            \If{$i$ in $f$}
                \State continue
            \Else
                \State $r_{i} = R_{S}[i]$
                \For{$p,\mu$ in $r_{i}$}
                    \If {$uncertainty = 0$ and $\mu.slide\_name$ not in $V$}
                    \State $R_{ret}.insert(\mu)$
                    \State $V.insert(\mu.slide\_name)$
                    \ElsIf{$uncertainty > 0$ and $\mu.hamming\_dist \leq \theta_{h^{\prime}}$ and $\mu.slide\_name$ not in $V$}
                        \State $R_{ret}.insert(\mu)$
                        \State $V.insert(\mu.slide\_name)$
                    \EndIf
                \EndFor
            \EndIf
        \EndFor
    \State $R_{ret} \gets \textproc{Sorting}(R_{ret})$
    \State \Return{$R_{ret}[0:K]$}
    \EndFunction
\end{algorithmic}
\end{algorithm}
\end{singlespace} 
\vspace{-4mm}

\textproc{Uncertainty-Cal} (\textbf{Algorithm~\ref{algo5}}) takes $R_{I}$ as the input and calculates the uncertainty for each $r_{i}$ by entropy. The lower the entropy, the less uncertain the mosaic and vice versa. The output is the entropy of $r_{i}$ along with records that summarize the diagnosis occurrences and hamming distance of each element in $r_{i}$. The disadvantage of counting the occurrences naively in the entropy calculation is that the most frequent diagnosis in the anatomic site dominates the result and therefore downplays the importance of others. We introduce a weighted occurrence approach to address this issue. The approach counts the diagnosis occurrences by considering the percentage of the diagnosis in the given site and the diagnosis position in the retrieval results. It calculates the weight of each diagnosis in the anatomic sites by the reciprocal of the number of diagnosis. We normalize the weights such that summation of them equals a constant $N$. A diagnosis's final occurrence in $r_{i}$ is the multiplication of the normalized weight of diagnosis and the inverse of position where the diagnosis appears in $r_{i}$. Therefore, the same diagnosis can have different weighted occurrence because of its position in $r_{i}$.  By doing so, the less frequent diagnosis and the one with lower hamming distance (\textit{i.e.}, diagnosis close to the front of the retrieval results) gain more importance in the ranking process. After this stage, we also summarized $R_{I}$ by three metadata $S_{lb}$, $S_{m}$, and $S_{l}$ to facilitate the subsequent processes, which are defined as
\vspace{-6mm}

\begin{itemize}
    \item $S_{lb}$: A nested hash table that stores the index of $r$ in $R_{I}$ as the key and its weighted diagnosis occurrences table as value.
    \vspace{-4mm}
    \item $S_{m}$: An array that stores tuples composed of the index $i$, the entropy, the hamming distance of all mosaics, and the total number of mosaics for each $r_{i}$ in $R_{I}$.
    \vspace{-4mm}
    \item $S_{l}$: An array that stores the total number of mosaics for each $r_{i}$ in $R_{I}$.
\end{itemize}
\begin{singlespace}
\begin{algorithm}[b!]
    \caption{Uncertainty Calculation} \label{algo5}
    \begin{algorithmic}
        \Function{Weighted-Uncertainty-Cal}{$r_{i}$, $D\_inv$}
            \State $label\_cnt, dist \gets \{\},\{\}$
            \For{$pos\_index, \mu$ in $r_{i}$}
                \State $label\_cnt[\mu.diagnosis]\gets label\_cnt[\mu.diagnosis] + D\_inv[\mu.diagnosis]\times \frac{1}{pos\_index}$
                \State $dist.insert(\mu.hamming\_dist)$
            \EndFor
            \For{$lb, cnt$ in $label\_cnt$}
            \If{$cnt < 1$} 
                \State $label\_cnt[lb]\gets 1$
            \EndIf
            \EndFor
            \State $Ent = \textproc{Entropy}(label\_cnt)$
            \State \Return{$Ent, label\_cnt, dist$}
    \EndFunction
\end{algorithmic}
\end{algorithm}
\end{singlespace}
\vspace{-6mm}

\textproc{Clean} (\textbf{Algorithm~\ref{algo6}}) aims to remove the outlier and the mosaics that are less similar to the query one in $R_{I}$. It takes summaries of mosaic $S_{m}$ and $S_{l}$ from the previous stage as input, removing $r$ whose result length $|r|$ is less than $5\%$ or greater than $95\%$ quantile. Besides, we take the average of the mean of hamming distance in top 5 mosaics for each $r \in R_{I}$ as a threshold denoted by $\theta_{h^{\prime}}$, using it to filter out $r$ whose mean of hamming distance in top 5 retrieval is greater than $\theta_{h^{\prime}}$. After cleaning the results, we sort it based on uncertainty calculated from \textproc{Uncertainty-Cal} in ascending order.
\begin{singlespace}
\begin{algorithm}[t!]
    \caption{Results Cleaning} \label{algo6}
    \begin{algorithmic}
        \Function{Clean}{$S_{m}, S_{l}$}
        \State //$S_{m}$:An array that stores tuples composed of the index $i$, the entropy, the hamming distance of all mosaics, and the total number of mosaics for each $r_{i}$ in $R_{I}$.
        \State //$S_{l}$:An array that stores the total number of mosaics for each $r_{i}$ in $R_{I}$.
        \State $tmp \gets \{\}$
        \State $m \gets 5$
        \State $l \gets 3$
        \State // When the unique results length is less than 3, we keep the original $S_{m}$.
        \If{\textproc{Unique}($S_{l}) \geq l$}
            \For{$res$ in $S_{m}$}
            \If{$res.r.len() \leq \textproc{Quantile}(S_{l}, 5\%)$ or $res.r.len() \geq \textproc{Quantile}(S_{l}, 95\%)$}
            \State del res
            \Else
                \State $tmp.insert(\textproc{Mean}(res.dist[0:m]))$
            \EndIf
            \EndFor
        \Else
            \For{$res$ in $S_{m}$}
                \State $tmp.insert(\textproc{Mean}(res.dist[0:m]))$
            \EndFor
        \EndIf
        \State $\theta_{h^{\prime}} \gets \textproc{Mean}(tmp)$
        \For{$res$ in $S_{m}$}
        \If{$\textproc{Mean}(res.dist[0:m]) > \theta_{h}$}
        \State del $res$
        \EndIf
        \EndFor
    \State $S_{m} \gets \textproc{Sort-Ascending}(S_{m}, key=Ent)$
    \State \Return{$S_{m}$}
    \EndFunction
\end{algorithmic}
\end{algorithm}
\end{singlespace} 
We can return the slide from $r_{i}$ in the front of the sorted $S_{m}$ based on the uncertainty However, the drawback is that the low uncertainty of first several $r \in S_{m}$ could be caused by the domination of the most frequent diagnosis in the given anatomic site. For example, the most frequent occurrences of the top 5 entries in $S_{m}$ could be KIRC, BLCA, KIPR, KIRP, and KIRP in the urinary site. In this case, the query slide should be better diagnosed as KIRP based on the majority vote. Therefore, the first and second entries that dominate the urinary site cases should not be considered during retrieval. We leverage the \textproc{Filtered-By-Prediction} (\textbf{Algorithm~\ref{algo7}}) to mitigate the issue. The function takes the summation of the diagnosis occurrences from the top 5 certain $s_{m}$ in $S_{m}$. It first uses the diagnosis with the maximum score as a pseudo ground truth diagnosis from top 5 certain $s_{m}$. Afterwards, it removes $s_{m}$ whose maximum occurrence diagnosis disagrees with the pesudo ground truth.

To return final results $R_{ret,I}$ of slide query $I$, we take slide name and its diagnosis in $r_{i}$ pointed by $S_{m}$ one by one. If the uncertainty of $r_{i}$ is zero, we take all $(p_{i},\mu_{i})$. On the contrary, we use $\theta_{h^{\prime}}$ again to ignore $(p_{i},\mu_{i})$ whose hamming distance is greater than the threshold. We sort $R_{ret,I}$ first by uncertainty in the ascending order then by the hamming distance in the descending order if the uncertainty is tie.

\begin{singlespace}
\begin{algorithm}[t!]
    \caption{Results Filtering by Prediction}\label{algo7}
    \begin{algorithmic}
        \Function{Filtered\_By\_Prediction}{$S_{m},S_{lb}$}
        \State//$S_{m}$: An array that stores tuples composed of the index $i$, the entropy, the hamming distance of all mosaics, and the total number of mosaics for each $r_{i}$ in $R_{I}$.
        \State//$S_{lb}$: A nested hash table that stores the index of $r$ in $R_{I}$ as the key and its weighted diagnosis occurrences table as value.
        \State $cnt \gets \{\}$
        \State $c \gets 5$
        \For{$s_{m}$ in  $S_{m}[0:c]$}
            \State // Calculate the score of each diagnosis
            \For{$d$ in $S_{lb}[s_{m}.i]$}
                \State $cnt[d] \gets cnt[d] + S_{lb}[s_{m}.i][d]$
            \EndFor
        \EndFor
        \State $plb\_list \gets \textproc{Sort-Descending}(cnt)$
        \State $p \gets 0$
        \State // A while loop is used here to avoid the case that the plb remove all $s_{m}$.
        \While{}
            \State $plb \gets plb\_list[p]$
            \State $removed \gets \{\}$
            \For{$s_{m}$ in $S_{m}[0:c]$}
                \State $pred \gets \textproc{Max}(S_{lb}[s_{m}.i])$
                \If{$pred \neq plb$}
                    \State $removed.insert(s_{m}.i)$
                \EndIf
            \EndFor
            \If{$removed.len() \neq c$}
                \State break
            \Else
                \State $p \gets p + 1$
            \EndIf
        \EndWhile
    \State \Return{$removed$}
    \EndFunction
\end{algorithmic}
\end{algorithm}
\end{singlespace}
\vspace{-6mm}
\noindent\textit{Training details of VQ-VAE} We used a sampled version of TCGA slide data in the first experiment (\textit{i.e.}, disease subtype retrieval in TCGA) to train our VQ-VAE. For each slide, we sample 10 1024x1024 patches at 20$\times$. All patches are converted from RGB to Pytorch tensor then normalized to $\left[-1, 1\right]$. The model was trained by Adam optimizer with a learning rate of $10^{-3}$ without weight decay and AMSgrad. We used default setting for other hyperparamters in Adam (\textit{i.e.}, $\beta s=(0.9, 0.999)$ and $\epsilon=10^{-8}$). We trained our model with a batch size of 4 for 10 epochs. We applied gradient clipping techniques by setting the gradient threshold to 1.0. The hyperparatmer $\beta$ in VQ-VAE was also set to 1.

\noindent\textbf{Ablation Study} We conducted ablation study on our ranking module to test the benefit of each function. Specifically, we compared the performance of following four settings: (1) Naive: removing \textproc{Clean} and \textproc{Filtered-By-Prediction} and treating each diagnosis occurrence in the mosaic retrieval result equally (\textit{i.e.,} replacing the assignment in line 4 in \textbf{Algorithm~\ref{algo5}} with 1.) (2)$+$Weighed count: applying \textproc{Uncertainty-Cal} to the ranking module only. (3)$+$Clean: applying \textproc{Uncertainty-Cal} and \textproc{Clean} to the ranking module. (4)$+$Filter: applying all functions to the ranking module.

\noindent\textbf{Visualization} We build confusion matrix for each site,  using each slide diagnosis as ground truth along the x-axis and $\textproc{Mv}(\textrm{ret}[:k])$ as predicted diagnosis along the y-axis. For the hamming distance matrix, we inspect the hamming distance between the query slide and each result in $\textrm{ret}[:k]$ one by one, adding the hamming distance to the associated diagnosis label and infinity to others. The infinity here is defined as hamming distance threshold $\theta_{h}$ plus 1 as $\theta_{h}$ is the maximum distance we can have in our pipeline. The final hamming distance matrix is obtained by dividing the total number of slides in the given anatomic site.

\noindent\textbf{Evaluation Metrics}\\
For all experiments, we remove slide with the same patient id as the query slide in the database (\textit{i.e.,} leave-one-patient-out evaluation). We use the mean majority vote ($\textproc{Mv}(\cdot)$) results in the top $k$ retrieval (mMV@k) instead of top-k accuracy over all instances in the data as this metric is more suitable to medical domain\cite{kalra2020pan}. We also use mean average precision at $k$ (mAP@k) to further evaluate the retrieval performance. Specifically,

\vspace{-12mm}
\begin{align}
    \textrm{mMV@k} &= \frac{1}{Q}\sum_{i=1}^{Q}\mathds{1}[L_{i} =\textproc{Mv}(\textrm{ret}_{i}[:k])]~,\\
    \textrm{mAP@k} &= \frac{1}{Q}~\sum_{i=1}^{Q}\frac{1}{M_{i} }\Bigg(\sum_{j=1}^{k}\textproc{Rel}(\textrm{ret}_{i}[j],L_{i})\frac{\sum_{m = 1}^{m = j}\textproc{Rel}(\textproc{Mv}(\textrm{ret}_{i}[m]), L_{i})}{j}\Bigg)~,
\end{align}
where $Q$ is the number of slide, $L_{i}$ is the ground truth diagnosis of slide $S_{i}$, and the $\textproc{MV}(\textrm{ret}_{i}[:k])$ is the predicted diagnosis of $S_{i}$ taken from the majority vote of top $k$ retrieval $\textrm{ret}_{i}$. $\textproc{Rel}(\cdot, \cdot)$ is another indicator function that output $1$ if the two inputs are the same and $0$ otherwise. We use $M_{i}$ to denote the the number of times when predicted diagnosis matches the ground truth among the top $k$ retrievals of $S_{i}$. Note that mAP@k is more lenient metric compared to mMV@k as a model can get $100\%$ mAP@k by put only one relevant slide at the first place among top k retrievals while mMV@k score is still zero in this case. Therefore, higher mMV@k is more important in our application but we still report mAP@k to quantify model's capability to give the relevant slide higher rank. To fairly compare with the best results in the paper\cite{kalra2020pan}. We set $k = 5$ for all our experiments except anatomic site retrieval, where $k$ is set to 10. In few cases, it is possible that the number of retrieval results is less than $k$. We consider a query is correct if the number of correct retrieval is greater than $\lceil \frac{k}{2} \rceil$.

\noindent\textbf{Computational Hardware and Software} \\
We stored all whole slide images (WSIs), patches, segmentation, mosaics across multiple disk with total size around 27T. Segmentation, patching mosaic extraction and search of WSIs were performed on CPU (AMD Ryzen Threadripper 3970X 32-Core Processor). The VQ-VAE pretraining and feature extraction were performed on 4 NVIDIA 2080Ti GPU. The whole FISH pipeline was written in Python (version 3.7.0) with the following external package: h5py (2.10.0), matplotlib(3.3.0), numpy (1.19.1), opencv-python (4.3.0.38), pillow (7.2.0), pandas (1.1.0), scikit-learn (0.23.1), seaborn (0.10), scikit-image (0.17.2), torchvision (0.6.0)  tensorboard (2.3.0) and tqdm (4.48.0). We used Pytorch (1.5.0) for deep learning. All plots were created by matplotlib (version 3.2.2) and seaborn (version 0.10.1). The internal function in Google Slide was used to plot pie chart.

\noindent\textbf{WSI dataset} \\
There are three dataset in our slide level retrieval experiment which are the diagnostic slide in The Cancer Genome Atlas (TCGA), Clinical Proteomic Tumor Analysis Consortium (CPTAC) and BWH in-house data.

\noindent\textit{TCGA diagnostic slide.} We downloaded all diagnostic slide from TCGA website. To fairly compare with Yottixel, we used slides from the same 13 anatomic sites for anatomic site retrieval and the same 29 diagnosis for disease subbtype retrieval. The detail slide and patient number are reported in \textbf{Extended Table~1}.

\noindent\textit{CPTAC diagnostic slide} 
We downloaded the tumor tissue slide of from the official website.  There are 503 CPTAC-CCRC slides from 216 patient, 544 CPTAC-UCEC slides from 240 patients, 679 CPTAC-LUSC slides from 210 patients, 669 CPTAC-LUAD slides from 224 patients and CPTAC-SKCM 283 slides from 93 patients. All slides are at 20$\times$.

\noindent\textit{BWH in house dataset} In this cohort, each whole slide image is from different patient. For the prostate data used in patch-level retrieval, we collected 23 slides at 20$\times$ and annotated regions in each slide by GP3, GP4, GP5, and normal. The detail slide and patient number are reported in \textbf{Extended Table~2}.

\noindent\textbf{WSI processing}\\
\noindent\textit{Segmentation.} We used the automatic segmentation tool in CLAM\cite{lu2021data} to generate the segmentation mask for each slide. The tool first applies binary threshold to a downsampled whole slide image on the HSV color space to generate a binary mask then refines the mask by median blurring and morphological closing to remove the artifacts. After getting the approximate contours of the tissue, the tool filters out the tissue contours and cavities based on certain area threshold.\\
\noindent\textit{Patching.} After segmentation, we cropped the contours into $1024\times1024$ patches without overlapping at 20$\times$. For 40$\times$ whole slide, we first cropped it into $2048\times2048$ patches then downsampled them into $1024\times1024$ to get the equivalent patches at 20$\times$.\\
\noindent\textit{Mosaic generation.} We followed the the mosaic generation process proposed in the paper\cite{kalra2020yottixel}. The algorithm first apply K-mean clustering on the RGB features extracted from each patches with number of cluster $K = 9$. Within each cluster, we run K-mean clustering again on the coordinate of each patch by setting the number of cluster equals to $5\%$ of the cluster size. If the number of cluster is less than 1 in the second stage, we took all coordinates within that cluster. Except the number of clusters, we used all default settings in Scikit-learn for K-mean clustering. To get better quality of mosaics, we collected 101 patches for both debris/pen smudges and tissue to train a logistic regression based on local binary pattern histogram feature to remove the unmeaningful regions. We used default setting from Scikit-learn package in logistic regression and used rotate invariant binary pattern from Scikit-image package with $P=8$ and $R=1$. The bin number of histogram was set to 128. \\
\noindent\textit{Artifacts removal.}  We found that there could be pure white mosaic in some rare cases after mosaic generation. We removed the such mosaic from the slide if the white region accounts for over $90\%$ of area. We applied binary threshold method in OpenCV with threshold value equals to 235 to determine the area of white regions.

\noindent\textbf{Data Availability} \\
The TCGA diagnostic whole slide image are available from TCGA website and CPTAC data is available from the NIH cancer imaging archive. The kather100k data are available from link provided in the paper\cite{kather2019predicting}. Reasonable requests for in-house BWH whole slide and prostate data may be addressed to the corresponding author.\\
\noindent\textbf{Code Availability}
We implemented all our methods in Python and using Pytorch as the primary package for training VQ-VAE. All scripts, checkpoints, preprocessed mosaics, and pre-built database to reproduce our experiment in the paper are available at \url{https://github.com/mahmoodlab/FISH}. All source code is licensed under GNU GPv3.\\
\noindent\textbf{Author Contributions}\\
C.C. and F.M. conceived the study and designed the experiments. C.C. performed the experiments. C.C. M.Y.L. D.F.K.W T.C. A.J.S. and F.M. analyzed the results. D.W. conducted the reader study. All authors wrote and approved the final paper. \\
\noindent\textbf{Acknowledgement}\\
This work was supported in part by internal funds from BWH Pathology, Google Cloud Research Grant and Nvidia GPU Grant Program and NIGMS R35GM138216 (F.M.). The content is solely the responsibility of the authors and does not reflect the official views of the National Institutes of Health, or the National Institute of General Medical Sciences.
\vspace{-6mm}
\section*{Competing Interests}
\vspace{-6mm}
The authors declare that they have no competing financial interests.
\vspace{-6mm}

\section*{Ethics Oversight}
\vspace{-6mm}
The study was approved by the Mass General Brigham (MGB) IRB office under protocol 2020P000233.

\end{spacing}

%\begin{nolinenumbers}
\vspace{-9mm}

\section*{References} 
\vspace{2mm}

\begin{spacing}{1.3}
\bibliographystyle{naturemag}

\end{spacing}
%\end{nolinenumbers}
\begin{figure*}[t!]
\centering
\includegraphics[width=0.9\textwidth]{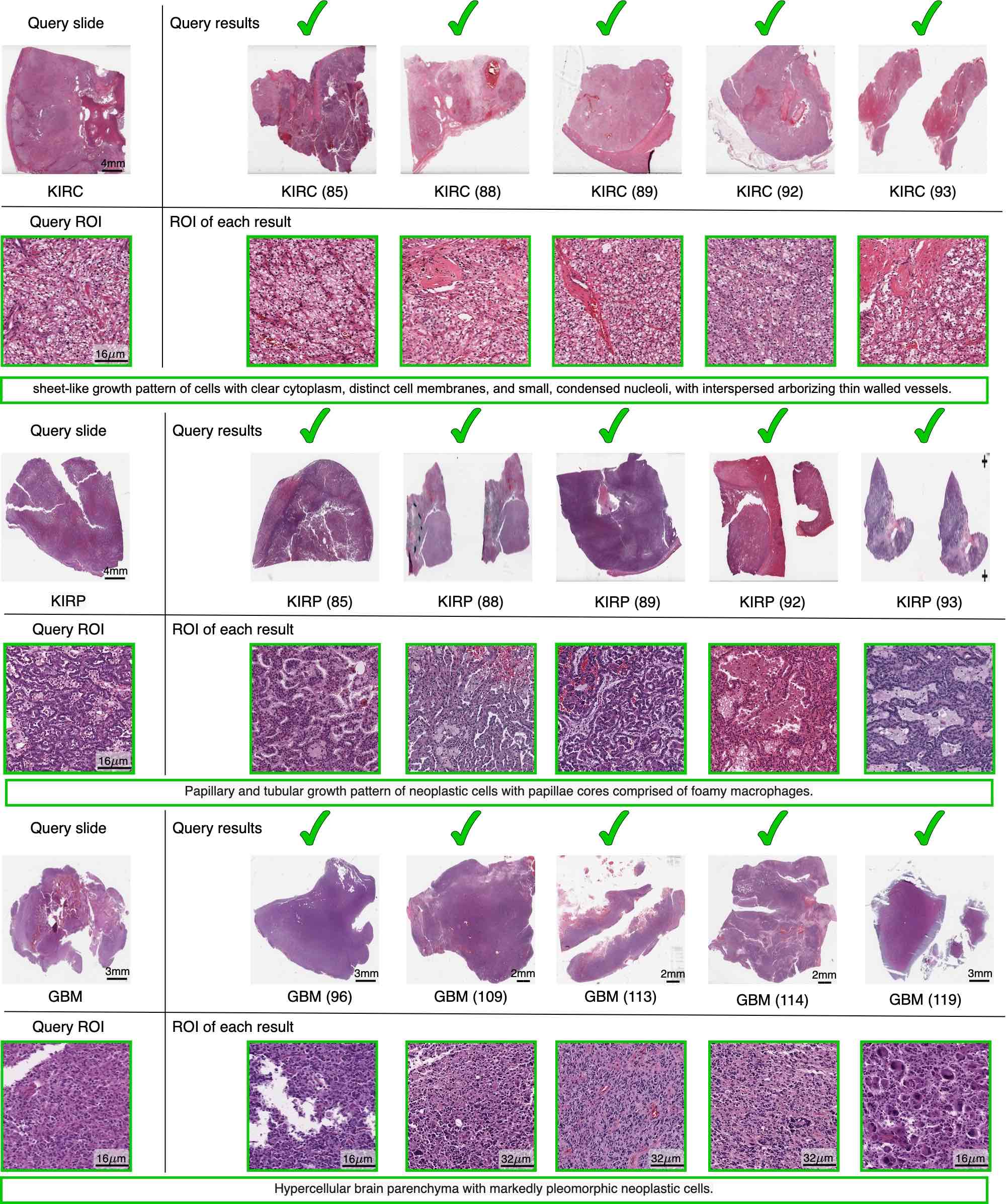}
\caption*{\textbf{Extended Data Figure 1. Examples of fixed-site disease subtype retrieval in TCGA cohort.} Examples of retrieval slides and corresponding ROI identified by FISH in TCGA-KIRC, TCGA-KIRP, and TCGA-GBM. The green border of ROIs denotes the selected regions match the histological features annotated by the pathologist. The number in parentheses is the hamming distance between the query slide and each result, determined by the identified ROI in each WSI. Each row shares the same scale bar unless specified otherwise.}
\label{fig_9}
\end{figure*}
\begin{figure*}[t!]
\centering
\includegraphics[width=0.9\textwidth]{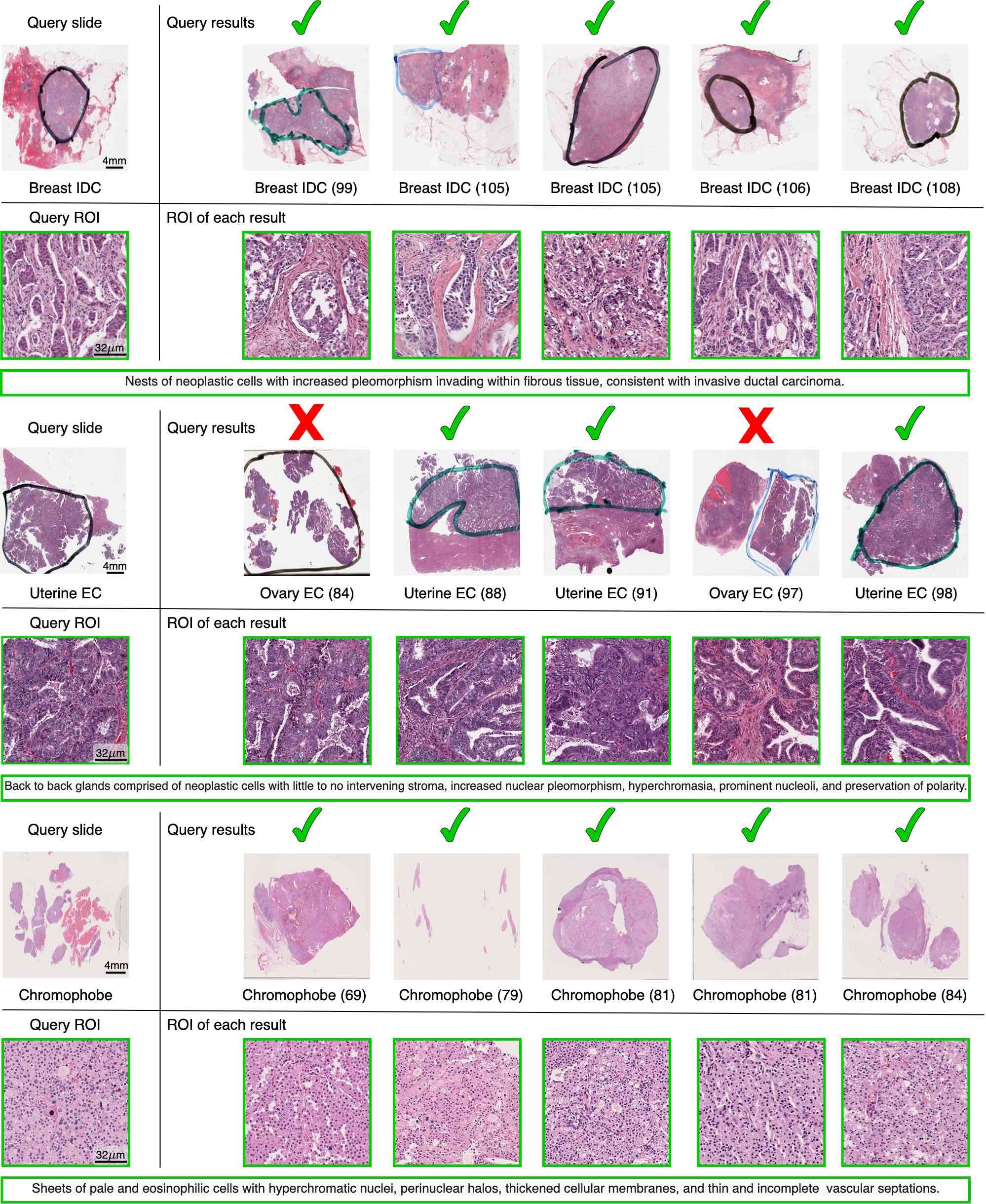}
\caption*{\textbf{Extended Data Figure 2. Examples of fixed-site disease subtype retrieval in BWH cohorts.} Examples of retrieval slides and corresponding ROI identified by FISH in Breast Invasive Ductial Carcinoma (Breast IDC), Uterine Endometriod Carcinoma (Uterine EC), and Kidney Chromophobe. The green border of ROIs denotes the selected regions match the histological features annotated by the pathologist. The number in parentheses is the hamming distance between the query slide and each result, determined by the identified ROI in each WSI. Each row shares the same scale bar unless specified otherwise. We found that FISH is sometimes confused with Ovary EC and Uterine EC, which is reasonable as both diseases have similar morphology.}
\end{figure*}
\begin{figure*}[t!]
\centering
\includegraphics[width=0.9\textwidth]{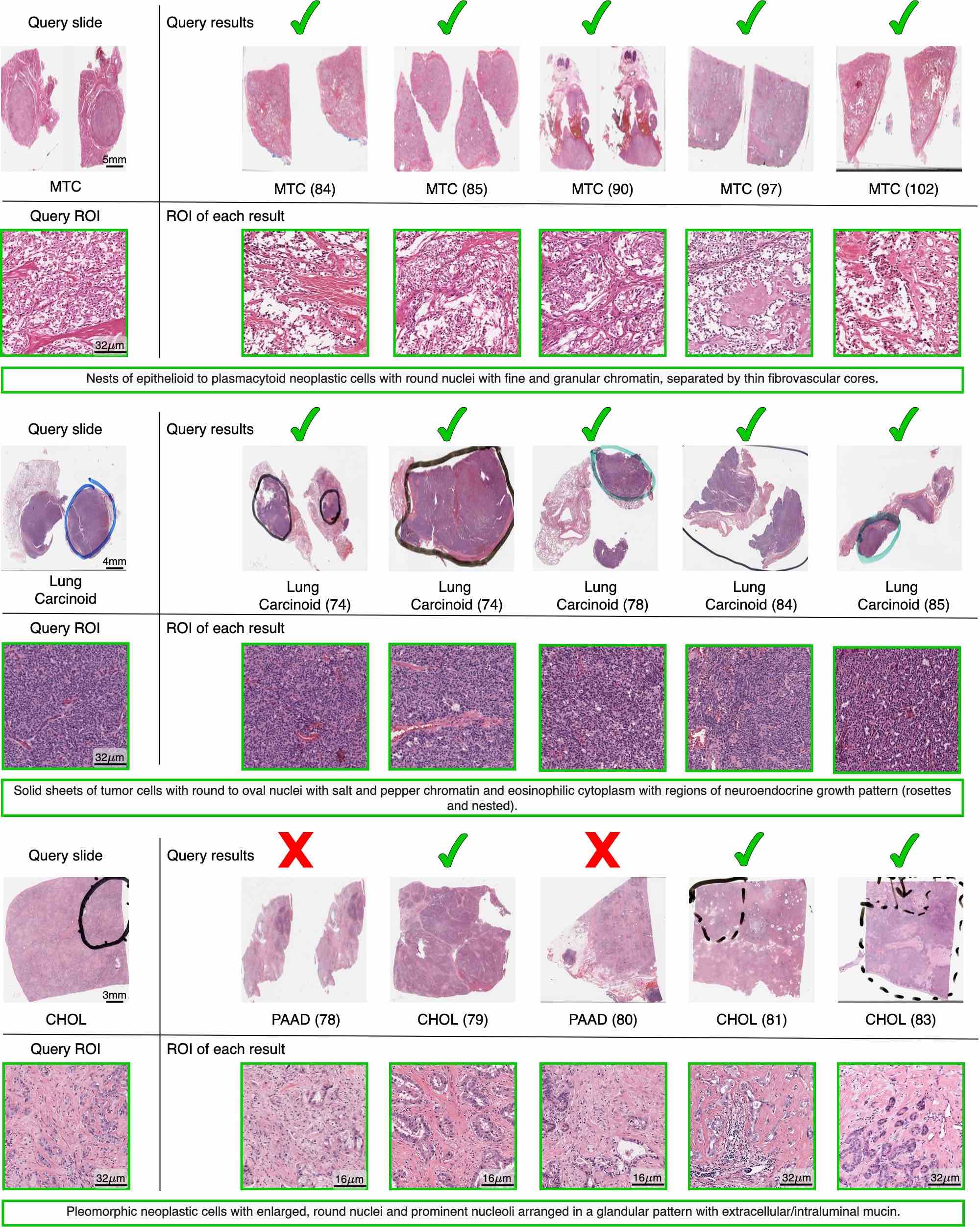}
\caption*{\textbf{Extended Data Figure 3. Examples of fixed-site retrieval on rare cancer subtype} Examples of retrieval slides and corresponding ROI identified by FISH in Medullary Thyroid Carcinoma (MTC), Lung Carcinoid and Cholangiocarinoma (CHOL). The green border of ROIs denotes the selected regions match the histological features annotated by the pathologist. The number in parentheses is the hamming distance between the query slide and each result, determined by the identified ROI in each WSI. Each row shares the same scale bar unless specified otherwise. We found that FISH is sometimes confused with Cholangiocarcinoma and Pancreatic Adenocarcinoma (PAAD), which is reasonable as both diseases have similar morphology.}
\end{figure*}
\begin{figure*}[t!]
\centering
\includegraphics[width=0.9\textwidth]{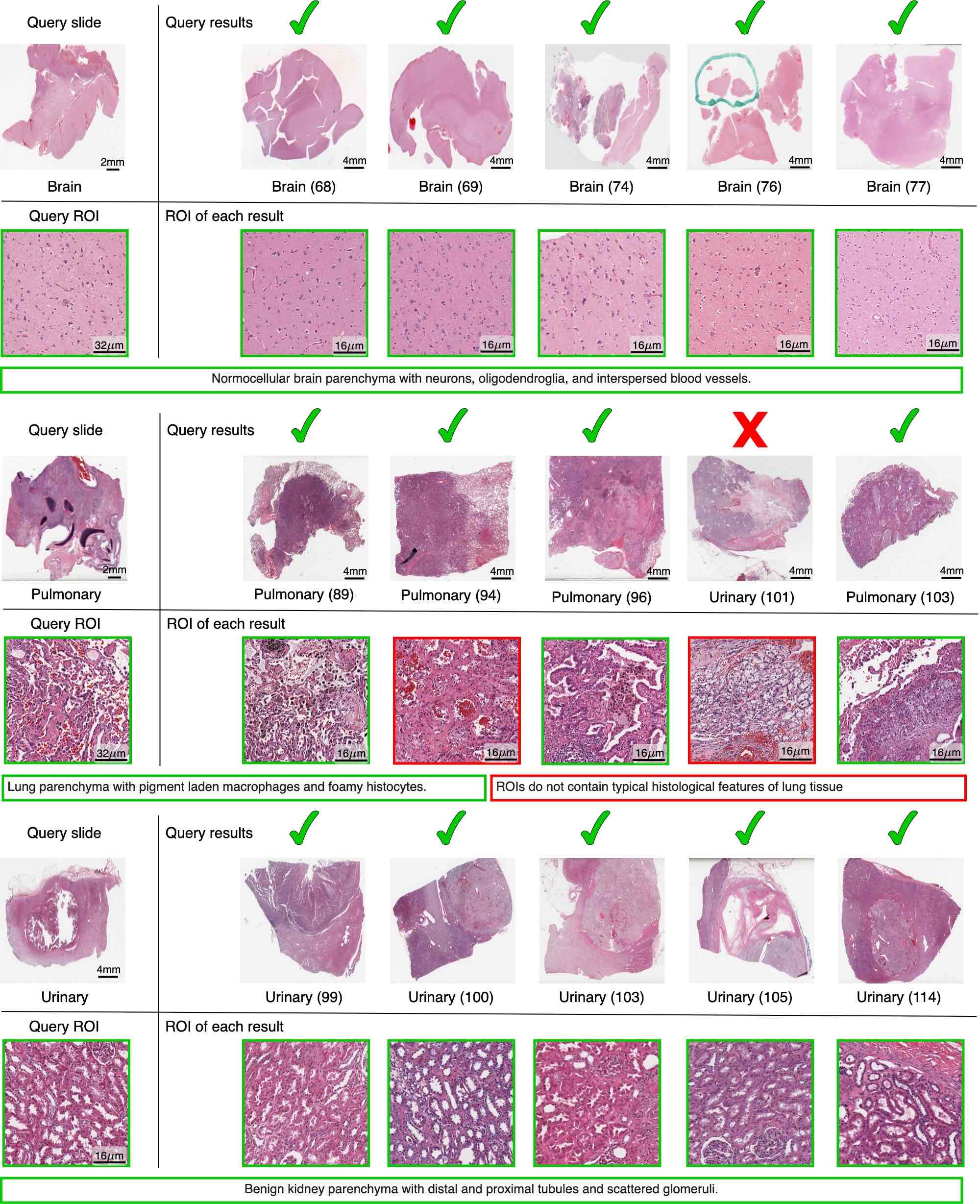}
\caption*{\textbf{Extended Data Figure 4. Examples of anatomic site retrieval in TCGA cohort.} Examples of retrieval slides and corresponding ROI identified by FISH in Brain, Pulmonary and Kidney. The visualization showed that FISH can also identified regions that contain typical histological features for a site. The green border denotes the regions that contain typical feautures while the red borders denote the failure cases. The number in parentheses is the hamming distance between the query slide and each result, determined by the identified ROI in each WSI. Each row shares the same scale bar unless specified otherwise.}
\end{figure*}
\begin{figure*}[t!]
\centering
\includegraphics[width=0.70\textwidth]{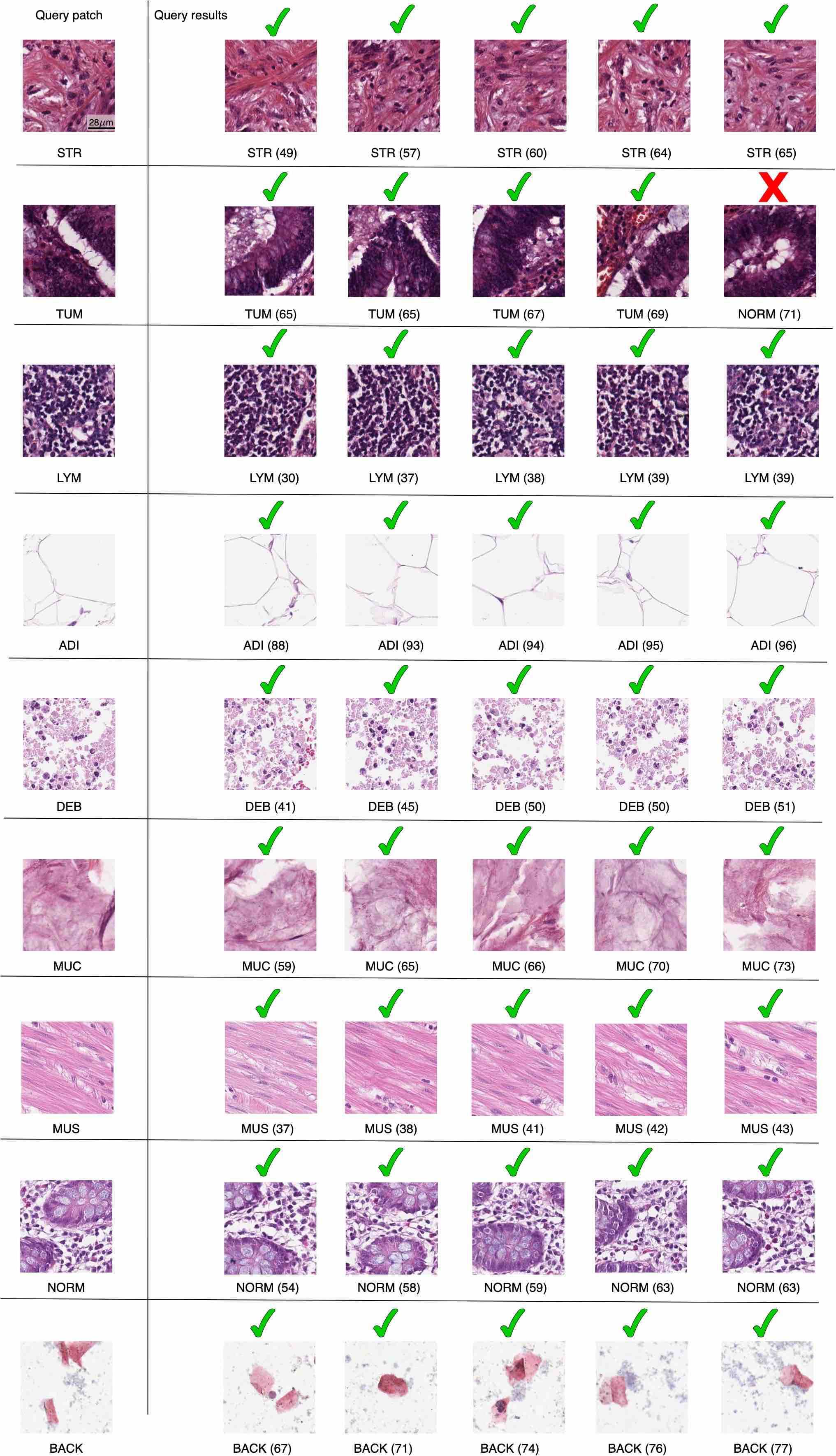}
\caption*{\textbf{Extended Data Figure 5. Examples of patch retrieval on Kather100k} The patches of cancer associated stroma (STR), colorectal adenocarcinoma epithelium (TUM), lymphocytes (LYM), adipose (ADI), debris (DEB), mucus (MUC), muscle (MUS), normal tissue (NORM), and background (BACK) are presented in the figure. The number in parentheses is the hamming distance between the query patch and each result. All patches share the same scale bar.}
\end{figure*}
\begin{figure*}[t!]
\centering
\includegraphics[width=0.85\textwidth]{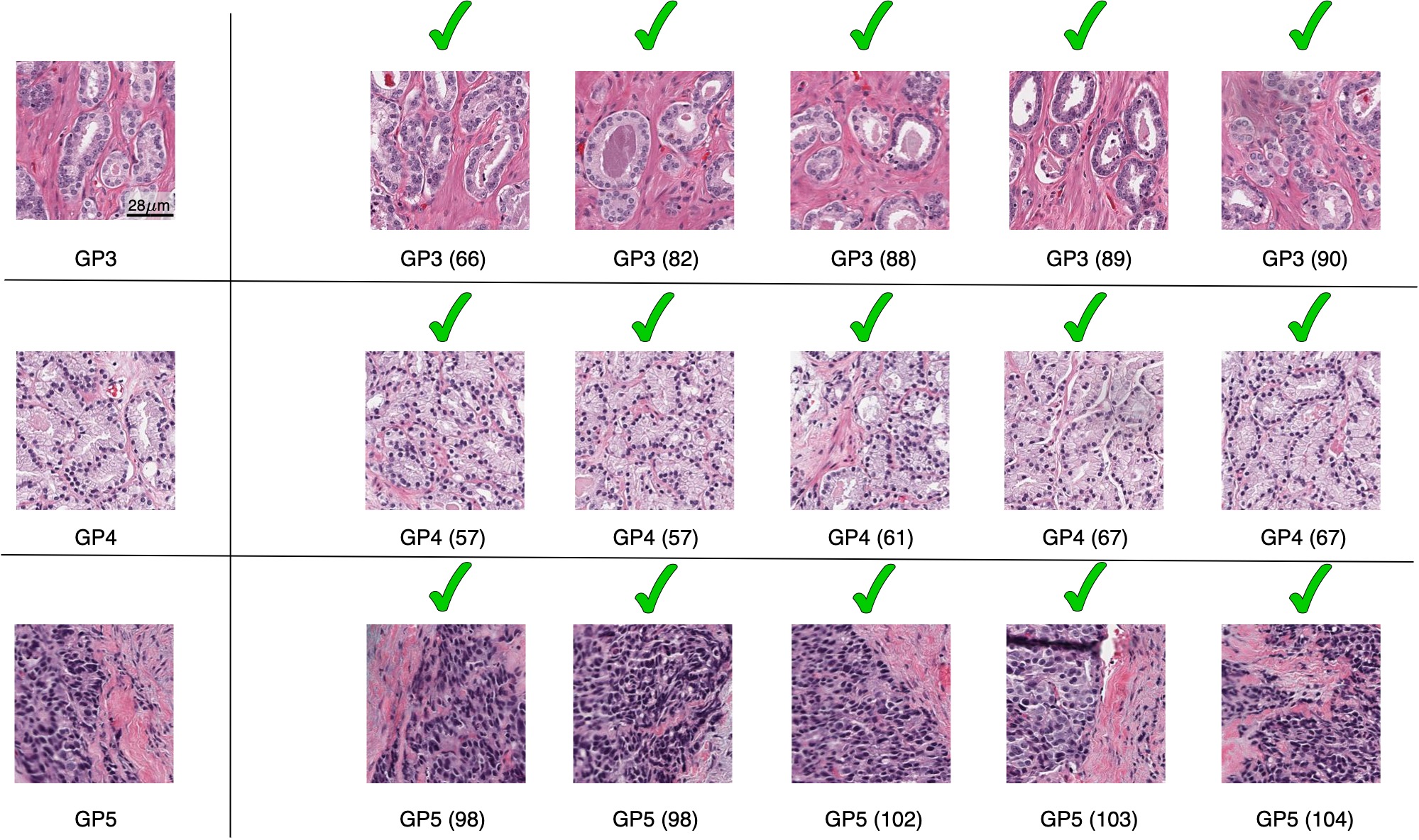}
\caption*{\textbf{Extended Data Figure 6. Examples of patch retrieval on in-house prostate data.} The patches of Gleason score 3, 4 and 5 are presented in the figure. The number in parentheses is the hamming distance between the query patch and each result. All patches share the same scale bar.}
\end{figure*}
\begin{figure*}[t!]
    \centering
    \includegraphics[width=0.9\textwidth]{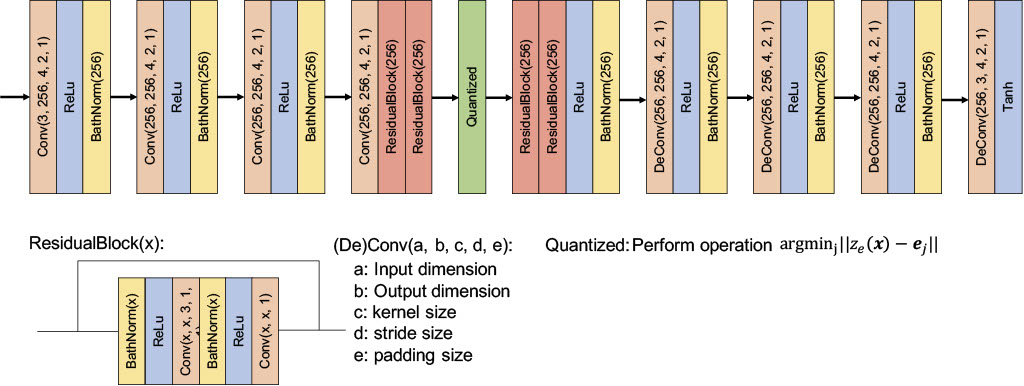}
    \caption*{\textbf{Extended Data Figure 7. VQ-VAE model architecture.} The VQ-VAE architecture we used in our pipeline. The modules before the Quantized layer are encoder made up by a sequence of convolution, relu activation, batch normalization and residual module. After quantized layer is a decoder made up by a sequence of residual module, deconvolution, batch normalization and relu activation. We applied a tanh activation for the output layer.} 
\end{figure*}

\begin{figure*}[b!]
    \centering
    \includegraphics[width=0.9\textwidth]{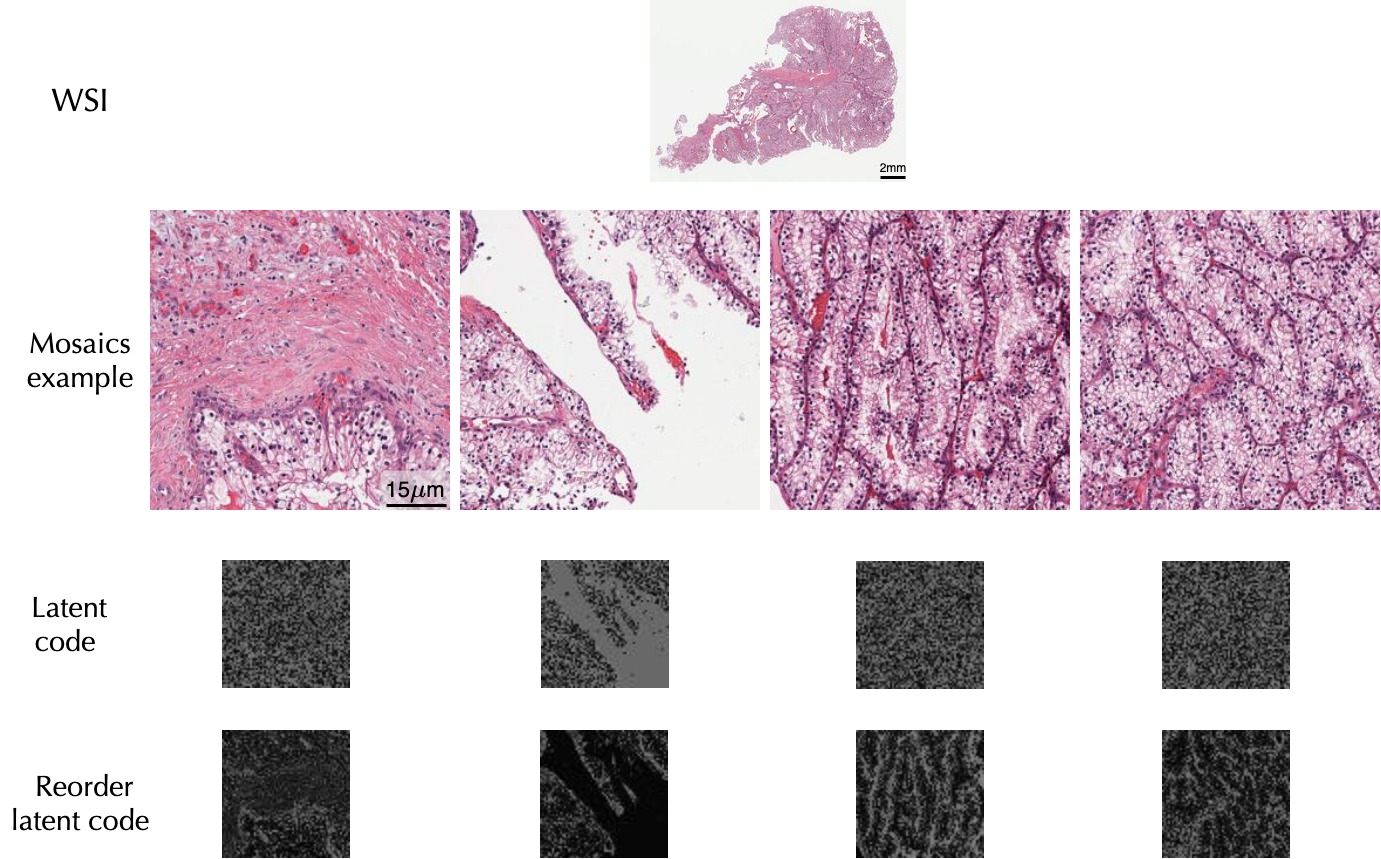}
    \caption*{\textbf{Extended Data Figure 8. Example of encoding a whole slide image.} An example of how we convert a WSI to mosaics and reordered latent codes through mosaic generation, VQ-VAE encoding and reorder the latent code.  Note that the reordered latent codes contain more semantic information in the original mosaics.} 
\end{figure*}
\begin{table*}[b!]
    \centering
        \centering
        \begin{tabular}{cccc}
             Diagnosis & WSIs & Patients & source\\
             \toprule
                GBM       &      816       &    378  & TCGA\\
                LGG       &      838       &    490  & TCGA\\
                ACC       &      227       &     56  & TCGA\\
                PCPG      &      196       &    176  & TCGA\\
                THCA      &      518       &    505  & TCGA\\
                COAD      &      441       &    433  & TCGA\\
                ESCA      &      158       &    156  & TCGA\\
                READ      &      158        &   157  & TCGA\\
                STAD      &      357       &    332  & TCGA\\
                UCEC      &      566       &    505 & TCGA\\
                CESC      &      278       &    268 & TCGA\\
                UCS       &       87        &    53 & TCGA\\
                OV        &      107         &  106 & TCGA\\
                DLBC      &       43        &    43 & TCGA\\
                THYM      &      180          & 121 & TCGA\\
                UVM      &        80        &    80 & TCGA\\
                SKCM      &      473         &  431 & TCGA \\
                CHOL      &       39          &  39 & TCGA\\
                LIHC      &      371        &   363 & TCGA \\
                PAAD      &      209          & 183 & TCGA\\
                LUAD      &      530          & 467 & TCGA \\
                MESO      &       86          &  74 & TCGA  \\
                LUSC      &      512          & 478 & TCGA  \\
                BLCA      &      457          & 386 & TCGA\\
                KIRC      &      519          & 513 & TCGA\\
                KICH      &      121          & 109 & TCGA\\
                KIRP      &      297          & 273 & TCGA\\
                TGCT      &      254          & 149 & TCGA \\
                PRAD      &      449          & 403 & TCGA\\
                BRCA    &     1,123          & 1,054& TCGA \\
                HNSCC &  471          &  449 & TCGA\\
                Mesenchymal &  600          & 254  & TCGA\\
                C-KIRC   &     503           & 216     & CPTAC\\
                C-UCEC   &     544           & 240    & CPTAC\\
                C-SKCM   &      283          &   93  & CPTAC\\
                C-LUSC   &      679          &   210  & CPTAC\\
                C-LUAD  &       669         &   224  & CPTAC\\
                \midrule
                Total &        14,239      &    10,467   &\\
            \bottomrule
        \end{tabular}
    \caption*{\textbf{Extended Data Table 1. Summary of public data.} Slide and patient number of public cohorts used in the experiment. We merge all diagnosis in the Breast, Head and Neck and Mesenchymal since we only use site information for the experiment.}
\end{table*}
\begin{table*}[b!]
    \centering
        \centering
        \begin{tabular}{ccc}
             Diagnosis & WSIs & Patients\\
             \toprule
                Glioblastoma Multiforme       &      380       &    380 \\
                Low-Grade Glioma, NOS       &      62       &    62  \\
                Astrocytoma       &      46       &     46  \\
                Anaplastic Astrocytoma      &      30       &    30\\
                Oligodendroglioma      &     28       &    28  \\
                Pilocytic Astrocytoma & 20 & 20\\
                Anaplastic Ologodendroglioma & 14 & 14\\
                Papillary Thyroid Cancer      &      316       &    202  \\
                Medullary Thyroid Cancer      &      202       &    202  \\
                Follicular Thyroid Cancer      &    150        &   150  \\
                Anaplastic Thyroid Cancer      &    114       &    114 \\
                Hurthle Cell Thyroid Cancer      &      56      &    56 \\
                Colorectal Adenocarcinoma      &      1,024       &    1,024\\
                Esophageal Adenocarcinoma       &       178        &    178 \\
                Esophageal Squamous Cell Carcinoma        &      41         &  41 \\
                Anal Squamous Cell Carcinoma      &       39        &    39 \\
                Uterine Endometrioid Carcinoma      &      480          & 480 \\
                High-Grade Serous Ovarian Cancer      &        242        &    242 \\
                Uterine Papillary Serous Carcinoma      &      157         &  157  \\
                Endometrioid Ovarian Cancer      &       64          &  64 \\
                Clear Cell Ovarian Cancer      &    48        &   48  \\
                Cholangiocarcinoma      &      55         & 55 \\
                Hepatocellular carcinoma      &      47          & 47  \\
                Gallbladder cancer      &       37          &  37   \\
                Pancreatic Neuroendocrine Tumor & 77 & 77\\
                Melanoma      &      197         & 197  \\
                Merkel Cell Carcinoma      &      75          & 75 \\
                Cutaneous Squamous Cell Carcinoma      &      38          & 38 \\
                Lung adenocarcinoma      &      1,377          & 1,377 \\
                Lung squamous cell carcinoma      &      392          & 392 \\
                Lung Carcinoid      &      53          & 53  \\
                Small Lung Cell Cancer      &      28          & 28 \\
                Bladder Urothelial Carcinoma    &   406          & 406 \\
                Kidney renal clear cell carcinoma &  271          &  271 \\
                Kidney renal papillary cell carcinoma &  96          & 96  \\
                Kidney Chromophobe   &     67           & 67    \\
                Upper tract Urothelial Carcinoma   &     47           & 47    \\
                Wilms Tumor   &      43          &   43  \\
                Breast Invasive Ductal Carcinoma   &      859          &   859  \\
                Breast Invasive Lobular Carcinoma  &       290         &   290  \\
                Prostate (patch level experiment)                           &       23          & 23\\
                \midrule
                Total               &     8,169      & 8,169\\
            \bottomrule
        \end{tabular}
    \caption*{\textbf{Extended Data Table 2. Summary of in-house BWH data} Slide and patient number of BWH cohorts used in the experiments.}
\end{table*}

\begin{table*}[t!]
    \caption*{\textbf{Supplementary Table 1. Individual retrieval result on TCGA cohort.} The individual retrieval result of FISH for each anatomic site on TCGA ($n=9,367$). Each row represents a query slide and the top-5 results with the corresponding hamming distance.\\
    \\
    \textit{This large table is available in the accompanying excel sheet.}}
\end{table*}
\begin{table*}[t!]
    \caption*{\textbf{Supplementary Table 2. Individual retrieval result on TCGA plus CPTAC cohort.} The individual retrieval result of FISH for each anatomic site on TCGA plus CPTAC ($n=6,791$). Each row represents a query slide and the top-5 results with the corresponding hamming distance.\\
    \\
    \textit{This large table is available in the accompanying excel sheet.}}
\end{table*}
\begin{table*}[t!]
    \caption*{\textbf{Supplementary Table 3. Individual retrieval result on BWH independent test cohort.} The individual retrieval result of FISH for each anatomic site on BWH test cohort ($n=8,035$). Each row represents a query slide and the top-5 result with the corresponding hamming distance.\\
    \\
    \textit{This large table is available in the accompanying excel sheet.}}
\end{table*}
\begin{table*}[t!]
    \caption*{\textbf{Supplementary Table 4. Individual retrieval result on rare diseases.} The individual retrieval result of FISH for each anatomic site on rare diseases ($n=1,785$). Each row represents a query slide and the top-5 result with the corresponding hamming distance.\\
    \\
    \textit{This large table is available in the accompanying excel sheet.}}
\end{table*}
\begin{table*}[t!]
    \caption*{\textbf{Supplementary Table 5. Individual retrieval result on anatomic site retrieval on TCGA cohort.} The individual retrieval result of FISH for each anatomic site on TCGA ($n=11,561$). Each row represents a query slide and the top-10 results with the corresponding hamming distance.\\
    \\
    \textit{This large table is available in the accompanying excel sheet.}}
\end{table*}
\begin{table*}[t!]
    \caption*{\textbf{Supplementary Table 6. Individual retrieval result on Kather100k} The individual retrieval result of FISH for Kather100k ($n=100,000$). Each row represents a query slide and the top-5 result with the corresponding hamming distance.\\
    \\
    \textit{This large table is available in the accompanying excel sheet.}}
\end{table*}
\begin{table*}[t!]
    \caption*{\textbf{Supplementary Table 7. Individual retrieval result on BWH prostate.} The individual retrieval result of FISH for BWH prostate ($n=6,779$). Each row represents a query slide and the top-5 result with the corresponding hamming distance.\\
    \\
    \textit{This large table is available in the accompanying excel sheet.}}
\end{table*}
\end{document}